\documentclass[preprint,12pt]{elsarticle}

\usepackage{amssymb}
\usepackage{graphicx}
\usepackage{subfig}
\usepackage{verbatim,xcolor,colortbl}
\usepackage{bm}
\usepackage{caption}
\usepackage{amsfonts}
\usepackage{amsmath}
\usepackage{mathtools}
\usepackage{float}
\usepackage{hyperref}
\usepackage{algorithm}
\usepackage{algpseudocode}
\usepackage{tablefootnote}
\usepackage{tikz}
\usepackage{footnote}
\makesavenoteenv{tabular}
\makesavenoteenv{table}
\usetikzlibrary{positioning,shapes,arrows.meta, backgrounds, fit}
\usepackage{setspace}
\usepackage{graphicx}
\usepackage{lscape}
\usepackage{csvsimple}
\usepackage{pgfplotstable}
\usepackage{booktabs}
\usepackage{siunitx}
\usepackage{multicol}
\definecolor{Blues1}{HTML}{eff3ff}
\definecolor{Blues2}{HTML}{bdd7e7}
\definecolor{Blues3}{HTML}{6baed6}
\definecolor{Blues4}{HTML}{2171b5}

\tikzstyle{decision} = [diamond, draw, fill=blue!20, 
text width=5em, text badly centered, node distance=3cm, inner sep=0pt]
\tikzstyle{block} = [rectangle, draw, fill=blue!10, 
text width=6em, text centered, rounded corners, minimum height=3em]
\tikzstyle{line} = [draw, -latex']
\tikzstyle{cloud} = [draw, ellipse,fill=red!20, node distance=3cm,
minimum height=2em]

\newcolumntype{R}[1]{>{\raggedleft\arraybackslash}p{#1\columnwidth}}

\journal{Neural Networks}

\begin{document}

\begin{frontmatter}



\title{Differential radial basis function network \\for sequence modelling}

\author{Kojo~Sarfo~Gyamfi\corref{cor1}\fnref{a1}}
\fntext[a1]{Present address: \\ Data Insights and Analytics, Loblaw Companies Limited, 1 President's Choice Circle, Brampton, ON L6Y 5S5, Canada}
\cortext[cor1]{Corresponding author}
\ead{kojo.gyamfi@loblaw.ca}
\author{James~Brusey}
\ead{james.brusey@coventry.ac.uk}
\author{Elena~Gaura}
\ead{elena.gaura@coventry.ac.uk}
\address{School of Computing, Electronics and Mathematics, Coventry University, CV1 5FB, Coventry, United Kingdom}
\date{\today}

\begin{abstract}
We propose a differential radial basis function (RBF) network termed RBF-DiffNet\textemdash whose hidden layer blocks are partial differential equations (PDEs) linear in terms of the RBF\textemdash to make the baseline RBF network robust to noise in sequential data. Assuming that the sequential data derives from the discretisation of the solution to an underlying PDE, the differential RBF network learns constant linear coefficients of the PDE, consequently regularising the RBF network by following modified backward-Euler updates. We experimentally validate the differential RBF network on the logistic map chaotic timeseries as well as on $30$ real-world timeseries provided by Walmart in the M5 forecasting competition. The proposed model is compared with the normalised and unnormalised RBF networks, ARIMA, and ensembles of multilayer perceptrons (MLPs) and recurrent networks with long short-term memory (LSTM) blocks. From the experimental results, RBF-DiffNet consistently shows a marked reduction over the baseline RBF network in terms of the prediction error (e.g., $26\%$ reduction in the root mean squared scaled error on the M5 dataset); RBF-DiffNet also shows a comparable performance to the LSTM ensemble at less than one-sixteenth the LSTM computational time. Our proposed network consequently enables more accurate predictions\textemdash in the presence of observational noise\textemdash in sequence modelling tasks such as timeseries forecasting that leverage the model interpretability, fast training, and function approximation properties of the RBF network.
\end{abstract}




\end{frontmatter}

\section{Introduction}\label{introduction}
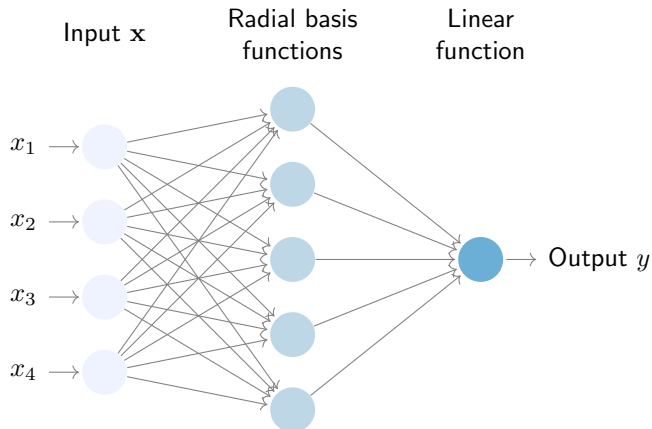
\begin{figure}[tbph]
    \centering
    \def\layersep{2.5cm}
{\sffamily \footnotesize
\begin{tikzpicture}[shorten >=1pt,->,draw=black!50, node distance=\layersep]
\tikzstyle{every pin edge}=[<-,shorten <=1pt]
\tikzstyle{neuron}=[circle,fill=black!25,minimum size=17pt,inner sep=0pt]
\tikzstyle{input neuron}=[neuron, fill=Blues1];
\tikzstyle{output neuron}=[neuron, fill=Blues3];
\tikzstyle{hidden neuron}=[neuron, fill=Blues2];
\tikzstyle{annot} = [text width=6em, text centered]

\foreach \name / \y in {1,...,4}
    \node[input neuron, pin=left:$x_{\y}$] (I-\name) at (0,-\y) {};

\foreach \name / \y in {1,...,5}
    \path[yshift=0.5cm]
        node[hidden neuron] (H-\name) at (\layersep,-\y cm) {};

\node[output neuron,pin={[pin edge={->}]right:Output $y$}, right of=H-3] (O) {};

\foreach \source in {1,...,4}
    \foreach \dest in {1,...,5}
        \path (I-\source) edge (H-\dest);

\foreach \source in {1,...,5}
    \path (H-\source) edge (O);

\node[annot,above of=H-1, node distance=1cm] (hl) {Radial basis functions};
\node[annot,left of=hl] {Input $\mathbf{x}$};
\node[annot,right of=hl] {Linear function};
\end{tikzpicture}
}
\caption{Radial basis function network (RBFN)}
\label{rbfnet}
\end{figure}

The radial basis function network (RBFN) is an artificial neural network first introduced by Broomhead and Lowe in the 1980s \cite{broomhead1988radial} but still very much in vogue now \cite{que2019back,masnadi2020attractor,dey2019robustness,teng2018machine} due to its robustness as a universal function approximator. Architecturally, it is a three-layer network having one input layer, one hidden layer, and one output layer, as shown in Figure \ref{rbfnet}, with activation units in the hidden layer made up of radial basis functions (RBFs).

The input layer is often connected to the hidden layer via direct connections whose weights are frozen at unity and thus not trainable, while the output layer is a simple linear layer. Though simple, this architecture is particularly efficient as a universal function approximator \cite{park1991universal,rizaner2018approximate,scarselli1998universal,girosi1990networks}; one reason for this is that the hidden layer of the RBF network performs a similar kernel transformation to an infinite-dimensional inner product space employed by other kernel machines such as the support vector machine \cite{abu2012learning,que2019back}. The RBF network has therefore seen many uses especially in timeseries forecasting, control and classification problems that occur in many real-world applications such as fraud detection, speech recognition, manufacturing, medical diagnosis, and face recognition \cite{dash2016radial,wong2011radial}. One other area in which the radial basis function network has seen increasing adoption is in the linear approximation of the value function in reinforcement learning in terms of the state-action variables \cite{sutton2018reinforcement,barreto2008restricted,kretchmar1997comparison}.

One reason for the ubiquity of RBF networks in many machine learning tasks is the interpretability of their outputs, since the hidden layer essentially performs a fuzzy nearest-neighbour association of an input vector to a set of well-defined examplars in the training data; thus, for a given input, the influence of different features on the output can be estimated from the relative importance of the features in these examplars. Another reason for the sustained use of the RBF network is the speed in training the network, since only the final linear layer is often trained; for the least-squares error (with ridge regularisation), there is, in fact, a closed-form solution for the network weights. This comes at the expense of a usually unsupervised step of selecting the number, centres and widths of the radial basis function activation units. Thus, the eventual performance of the RBF network is highly susceptible to the widths and centre locations of the RBF units \cite{lim2019distance,iske2000optimal,orr1995regularization,scheibel1999centre}, which in turn can be heavily influenced by the presence of noise in the data \cite{dey2019robustness,masnadi2020attractor}. Other neural network architectures may not be so exceptionally sensitive to noise. Several other approaches mainly using forward selection \cite{chen2006kernel,chen2008construction,gomm2000selecting} and sophisticated clustering methodologies or the self-organising map \cite{dash2016radial,huilan2005self,kamalabady2008new,lim2019distance} have thus been proposed to better locate the RBF centres.

Crucially however, for sequence modelling tasks such as timeseries forecasting, the sequence data has to be reshaped such that the RBF network takes as inputs the $l$ lagged values of the sequence (in addition to any exogenous inputs) and outputs some next points in the sequence; therefore, if there is a single noisy observation in the sequence, this observation likely gets replicated $l$ times in the reshaped data that is input to the RBF network. In the extreme case, the sequence data may be so corrupted that the data contains no more meaningful information for prediction \cite{masnadi2020attractor}. This profoundly degrades the optimal placement of the RBF centres and consequently the performance of the network. It is worth noting that this problem of noise propagation, as described, is not prevalent in the application of the RBF network to other regression or classification tasks where there are no temporal correlations in the data, in which case each noisy observation occurs only once in training. For example, in reinforcement learning settings, since the state transition in the sequential decision process is typically considered Markovian, there is functional dependence on only the last state vector, i.e. $l=1$, and thus a single noisy observation does not replicate itself in the reshaped training data.

In this paper, we propose a novel neural network architecture, termed the differential RBF network (RBF-DiffNet), that is designed to learn a representation of the sequence data that ignores signal noise. By utilising activation blocks made up of partial differential equations (PDEs) linear in terms of the radial basis function \textemdash with the constant linear coefficients of the PDE being trainable \textemdash we introduce a regularisation based on backward Euler updates that makes the network robust to noise. The intuition behind this contribution stems from the observation that many real-world sequence data derive from phenomena (such as in biology, economics or physics) whose underlying dynamics, in the absence of noise or control inputs, may be modelled by a set of partial differential equations, however complex. For example, the sequence of air temperature data observed in a car cabin may very well be described by a set of heat balance equations \cite{brusey2018reinforcement}, which are PDEs.

Our main contribution in this paper is therefore as follows: we introduce the differential RBF network and analyse its mathematical properties that enable the network to show robustness to noisy perturbations to sequential data in applications such as timeseries forecasting, where the baseline RBF network would be rather susceptible to the noisy training inputs.

This proposed network is detailed in Section 3. Section 4 presents an experimental validation of the proposed architecture on the logistic map chaotic timeseries \cite{maathuis2017predicting,farmer1987predicting} and $30$ different real-world retail timeseries from the M5 competition \cite{makridakis2020m5}; this section also includes performance comparisons with the baseline RBF network, autoregressive integrated moving average (ARIMA) model, ensembles of MLPs and recurrent neural networks with long-short-term memory (LSTM) blocks \cite{makridakis2020m5}. Conclusions are given in Section 5, while the problem statement and related work are presented in the next section.

\section{Problem statement and related work}\label{background}
We begin by considering a set of input-output pairs $\{\textbf{x}_n, y_n\}_{1:N}$ from which we wish to train a function $f: \mathcal{X} \to \mathcal{Y}$, where $\textbf{x}_n \in \mathcal{X}$ is a $d$-dimensional input vector and $y_n \in \mathcal{Y}$ is a scalar-valued output. For a sequence data $\{s_{\pi}\}_{1:T}$ of length $T$, we assume that this is reshaped into input-output pairs $\{\textbf{x}_n, y_n\}_{1:N}$ as before, where $\textbf{x}_n$ is given by the last $l$ realisations of the sequence prior to timestep $t_{\pi+1}$, i.e., $\textbf{x}_n^\top = [s_{\pi-l+1}, ..., s_\pi]$, and $s_{\pi+1} = f(\textbf{x}_n)$ where $y_n$ is the true output. If there are other exogenous inputs $\textbf{e}$ on which the sequence $\{s_{\pi}\}_{1:T}$ has dependence, we assume this is again captured in $\textbf{x}_n$ as $\textbf{x}_n^\top = [\textbf{e}^\top, s_{\pi-l+1}, ..., s_\pi]$. Note that if there are exogenous inputs, then $l< d$; otherwise $l = d$.

The radial basis function network (RBFN) as shown in fig. \ref{rbfnet}, can be written concisely as:
\begin{equation}\label{rbfn}
    f(\textbf{x}_n) = \sum_{j=1}^{c}w_j\phi_j(\textbf{x}_n)+w_0,
\end{equation}
where $c$ is the number of RBF centres, corresponding to the number of neurons in the hidden layer, and
\begin{equation}\label{rbf}
    \phi_j(\textbf{x}_n) = e^{-\beta_j\|\textbf{x}_n-\bm{\mu}_j\|^2}
\end{equation}
is the Gaussian radial basis function, which is the RBF we consider in this paper; the choice of non-linearity in other radial basis functions has been found not to be too crucial to the performance of the network \cite{chen1991orthogonal}. In \eqref{rbf}, $\bm{\mu}_j$ are examplars in the training data; $\beta_j$, as defined in \eqref{beta_eqn}, is inversely proportional to the width of the $j$th RBF, and $w_j, w_0$ are the RBF weights and bias to be optimised. 

From \eqref{rbfn} and \eqref{rbf}, the hidden layer of the RBF network essentially performs a fuzzy nearest-neighbour association of the input vector $\textbf{x}_n$ to the set of examplars $\bm{\mu}_j$, based on a Mahalanobis distance criterion with spherical covariance in terms of $\beta_j$. Thus, for any given input vector, the influence of different features on the output can be estimated from the relative importance of the features in each of the $c$ examplars, each weighted by its corresponding RBF weight $w_j$; this property makes the output of the RBF network interpretable in terms of its inputs.

For the sake of brevity, we drop the subscript $n$ in $\textbf{x}_n$ in the following. In matrix form, \eqref{rbfn} is equivalent to:
\begin{equation}\label{rbfn2}
    f(\textbf{x}) = \bm{\phi}(\textbf{x})^\top\tilde{\textbf{w}},
\end{equation}
where,
\begin{equation}
    \bm{\phi}(\textbf{x}) = [1, \phi_1(\textbf{x}), ..., \phi_c(\textbf{x})]^\top,
\end{equation}
and
\begin{equation}\label{w_w_tilde}
\tilde{\textbf{w}} = [w_0, w_1, ..., w_c]^\top
\end{equation}

If one considers the entire training set, then we have the following system of equations:
\begin{equation}
    \textbf{f}= \bm{\Phi}\tilde{\textbf{w}},
\end{equation}
where
\begin{equation}
    \textbf{f}= [f(\textbf{x}_1), f(\textbf{x}_2), ..., f(\textbf{x}_N)]^\top,
\end{equation}
and
\begin{equation}
    \bm{\Phi} = 
    \begin{bmatrix}
    1 & \phi_1(\textbf{x}_1) & \cdots & \phi_c(\textbf{x}_1)\\
    \vdots &\ddots&\ddots&\vdots \\
    1 & \phi_1(\textbf{x}_N) & \cdots & \phi_c(\textbf{x}_N)\\
    \end{bmatrix}
\end{equation}
For the least-squares error and low to moderate number of RBF nodes, $\tilde{\textbf{w}}$ can be optimised in closed-form as:
\begin{equation}\label{w_opt}
    \tilde{\textbf{w}}^* = (\bm{\Phi}^\top\bm{\Phi} + \gamma \textbf{I})^{-1}\bm{\Phi}\textbf{y},
\end{equation}
where $\gamma$ is a regularisation coefficient, $\textbf{I}$ is the identity matrix of size $c+1$, and $\textbf{y}^\top = [y_1, ..., y_n]$. In general, the weights $\tilde{\textbf{w}}$ can be trained for any arbitrary loss function using different optimisation routines.

Separate from the RBF network weights are the hyperparameters $c$, $\bm{\mu}_j$ and $\beta_j$ that require tuning. Most commonly, these are obtained from an unsupervised preprocessing step; the number of RBF centres is fixed and the examplars $\bm{\mu}_j$ are determined as the cluster centres obtained from some variants of K-Means clustering \cite{lim2019distance,scholkopf1997comparing,que2019back,masnadi2020attractor}, while $\beta_j$ is derived from the compactness of the individual clusters. Specifically, $\beta_j$ is defined as:
\begin{equation}\label{beta_eqn}
    \beta_j = \frac{1}{2\sigma_j^2},
\end{equation}
where popular choices of $\sigma_j$ \cite{mccormick2013radial,wu2012using,benoudjit2003kernel,moody1989fast} include:
\begin{align}
    &\sigma_j \coloneqq \frac{d_{max}}{\sqrt{2c}}, \quad \forall j\in\{1,...,c\} \\
    &\sigma_j \coloneqq \frac{1}{n_j}\sum_{\textbf{x}\in \mathcal{C}_j}\|\textbf{x}-\bm{\mu}_j\|, \quad \forall j\in\{1,...,c\}\label{rbf_width}\\
    &\sigma_j \coloneqq \frac{1}{r}\sum_{\textbf{x}\in \mathcal{R}_j}\|\textbf{x}-\bm{\mu}_j\|, \quad \forall j\in\{1,...,c\}\\
    & \sigma_j \coloneqq \frac{1}{cr}\sum_{j=1}^{c}\sum_{\textbf{x}\in \mathcal{R}_j}\|\textbf{x}-\bm{\mu}_j\|, \quad \forall j\in\{1,...,c\},
\end{align}
where $d_{max}$ is the maximum distance between the centres of any two clusters, $\mathcal{C}_j$ is the set of all points in the $j$th cluster, $n_j$ is the number of points in the $j$th cluster, and $\mathcal{R}_j$ is the set of the $r$ closest points to the $j$th cluster centre.
An alternative to K-Means clustering employed for RBF networks is the self-organising map which allows for a more intuitive determination of the number of cluster centres $c$ when the data is projected onto two or three dimensions \cite{lin2005time,kamalabady2008new}.

Since the RBF network parameters affect the network's performance quite significantly, other approaches utilising other performance metrics have been proposed for the selection of the hyperparameters. For example, the RBF nodes may be incrementally added in order to maximise the Fisher class-separability ratio and the leave-one-out cross validation RMSE for classification and regression tasks respectively \cite{chen2006kernel,chen2008construction}. Nevertheless, the presence of observational noise tends to influence the optimal placement of centres \cite{dey2019robustness} or any metrics computed based on them, such as the Fisher class-separability ratio. The prevalence of observational noise tends to have an even more profound effect for sequence data \cite{masnadi2020attractor}, where upon reshaping into input-output pairs, noisy observations get replicated throughout the data if the lookback window $l$ is much greater than $1$.

One way to make the RBF network robust is to normalise the network to achieve the so-called \textit{partition of unity} property where the $c$ normalised radial basis functions sum up to one for every point in the input space $\mathcal{X}$, making the RBF network less susceptible to noisy observations in arbitrary regions of the input space \cite{shorten1994normalising,shorten1996side}. Normalisation thus results in the RBF network losing its local characteristics, but often results in improved generalisability \cite{bugmann1998normalized}. The normalised RBF network $f_{norm}$ is given mathematically as:
\begin{equation}\label{nrbfn}
    f_{norm}(\textbf{x}_n) = \frac{\sum_{j=1}^{c}w_j\phi_j(\textbf{x}_n)}{\sum_{j=1}^{c}\phi_j(\textbf{x}_n)}+w_0,
\end{equation}

Since normalising the RBF network has several known side effects such as shifts in the maxima of the RBFs \cite{shorten1994normalising, shorten1996side}, in this paper, we take a different perspective towards making the RBF network robust specifically for its application to modelling sequential data. This involves utilising activation blocks made up of partial differential equations linear in terms of the radial basis function and based on backward Euler discretisation of the sequence data.

It is worth mentioning that our proposed differential RBF network architecture, although it utilises a similar Euler discretisation as the neural ordinary differential equation (ODE-Net) \cite{chen2018neural}, differs from the ODE-Net as follows: while the ODE-Net parameterises the derivatives of the hidden state of a residual network with another neural network in the limit of a large number of hidden layers when the discrete step size approaches $0$, the differential RBF network parameterises the solution to the differential equation with an RBF network and directly evaluates the derivatives of this solution according to the backward Euler updates given in Section \ref{proposed_section} in such a way as to regularise the original RBF network.

Furthermore, our work differs from other works that have employed the RBF network in solving ordinary or partial differential equations \cite{chen2016reduced,larsson2017least,mai2001numerical,lagaris1998artificial,schaback1970using} in that, instead of a well-defined set of differential equations to solve, we have only some discrete realisations from phenomena that are assumed to be described by unknown differential equations. Thus, we start with a solution to an unknown differential equation in the form of the RBF network and work backwards to find an optimal differential equation with constant coefficients that best describes the sequence data, based on backward Euler updates.

\section{Proposed differential RBF network}\label{proposed_section}
\subsection{Intuition}
We first consider the univariate series $\{s_{\pi}\}_{1:T}$, and assume that this sequence data is generated by some underlying differential equation given by:
\begin{equation}
    \frac{dz}{dt} = g(t),
\end{equation}
that is, the sequence $\{s_t\}_{1:T}$ comprises discrete realisations of this underlying process. These discrete realisations may be approximated by backward Euler updates \cite{atkinson2011numerical} of the form:
\begin{equation}\label{euler}
    s_{\pi+1} = s_{\pi} + hg(t_{\pi+1}, s_{\pi+1}),
\end{equation}
where $h$ is some small interval between the occurrences of $s_{\pi}$ and $s_{\pi+1}$, i.e., $h = t_{\pi+1}-t_{\pi}$. Note that in a noiseless system, $s_{\pi} = z(t_{\pi})$.
The backward Euler update in \eqref{euler} can be thought of as a first-order Taylor's approximation, and thus to improve this approximation and reduce the truncation errors, we may consider a higher-order Taylor's expansion up to degree $\nu$  as follows:
\begin{equation}
    s_{\pi+1} = s_{\pi} + hz^{\prime}(t_{\pi+1}, s_{\pi+1}) + ... + \frac{h^\nu}{\nu!}z^{(\nu)}(t_{\pi+1}, s_{\pi+1}).
\end{equation}
However, since we wish to predict the next state of the sequence based on its prior values, we replace the functional dependence of $z$ on time $t_{\pi+1}$ and $s_{\pi+1}$ with the last $l$ realisations of the series, obtaining an analogous form, using multi-index notation as:
\begin{align}\label{higher}
    &s_{\pi+1} = \bm{\lambda}^\top\textbf{s}_{\pi-l+1:\pi} + \textbf{h}Dz(s_{\pi-l+1},..,s_{\pi}) + \frac{\textbf{h}^2}{2}D^2z(s_{\pi-l+1},..,s_{\pi}) + ... \nonumber\\
    &+ \frac{\textbf{h}^\nu}{\nu!}D^{(\nu)}z(s_{\pi-l+1},..,s_{\pi}),
\end{align}
where $\textbf{h}$ is now a small vector interval, $D^{i}z$ is an $i$th-order tensor, with $Dz$ and $D^2z$ being the gradient and Hessian respectively of $z$, and $\bm{\lambda}^\top\textbf{s}_{\pi-l+1:\pi}$ is the weighted sum of the last $l$ realisations of the sequence, with $\bm{\lambda}$ being the vector of weights.

We may then generalise \eqref{higher} to consider a multivariate series as follows:
\begin{equation}\label{higher2}
    s_{\pi+1} = \bm{\lambda}^\top\textbf{s}_{\pi-l+1:\pi} + \textbf{h}Dz(\textbf{x}) + \frac{\textbf{h}^2}{2}D^2z(\textbf{x}) + ... + \frac{\textbf{h}^\nu}{\nu!}D^{(\nu)}z(\textbf{x}).
\end{equation}
Here, as mentioned previously in Section \ref{background}, $\textbf{x}$ encapsulates the lagged inputs of the series as well as exogenous inputs, and $s_{\pi+1} = z(\textbf{x}) \in \mathcal{Y}$.
For the special case where $\nu = 2$, since $Dz$ and $D^2z$ are respectively the gradient and Hessian of $z$, we have from \eqref{higher2} that,
\begin{equation}\label{nu=2}
    s_{\pi+1} = \bm{\lambda}^\top\textbf{s}_{\pi-l+1:\pi} + \sum_{i=1}^{d}h_i\frac{\partial z(\textbf{x})}{\partial x_i} + \sum_{i=1}^{d}\sum_{p=1}^{d}h_i h_p \frac{\partial^{2}z(\textbf{x})}{\partial x_i x_p},
\end{equation}
where $h_i, h_p$ are respectively the $i$th and $p$th components of $\textbf{h}$.

To keep the computation in \eqref{higher2} tractable, we ignore all mixed derivatives (such as $\frac{\partial^2 z(\textbf{x})}{\partial x_i \partial x_p}$ in the summation in \eqref{nu=2} involving the second-order tensor, where $i\neq p$) thus approximating \eqref{higher2} with the following partial differential equation:
\begin{equation}\label{pde}
    s_{\pi+1} \approx \bm{\lambda}^\top\textbf{s}_{\pi-l+1:\pi} + \sum_{k=1}^{\nu}\sum_{i=1}^{d}a_{k,i}\frac{\partial^{(k)}z(\textbf{x})}{\partial x_i^k},
\end{equation}
where the coefficients $a_{k,i}$ (which replace $\textbf{h}$ in \eqref{higher2}) now have to be optimised to maximise the fit to the data.

Since $z: \mathcal{X} \to \mathcal{Y}$, where $\textbf{x} \in \mathcal{X}$ and $s_{\pi+1} = z(\textbf{x}) \in \mathcal{Y}$, a good choice for $z$ is the radial basis function network $f$ defined in \eqref{rbfn}, i.e., 
\begin{equation}\label{rbfn3}
    z(\textbf{x}) \coloneqq f(\textbf{x}) = \sum_{j=1}^{c}w_j\phi_j(\textbf{x})+w_0.
\end{equation}
With $z$ defined as above, we have from \eqref{pde} that:
\begin{equation}\label{pde2}
    f(\textbf{x}) = \bm{\lambda}^\top\textbf{s}_{\pi-l+1:\pi} + \sum_{k=1}^{\nu}\sum_{i=1}^{d}a_{k,i}\frac{\partial^{(k)}f(\textbf{x})}{\partial x_i^k}.
\end{equation}
The utility in defining the function $z$ as a radial basis function network $f$ is in its parameter efficiency, i.e., we are able to obtain closed-form expressions for its higher-order derivatives without any increase in the number of variables that parameterise $z$ or these higher derivatives. 

The relationship in \eqref{pde2} then represents a regularisation of the parameters of $f$ in \eqref{rbfn}. Specifically, while the radial basis function network $f$ is ordinarily given by \eqref{rbfn}, we are now subjecting it to the additional constraint that the next observation $s_{\pi+1}=f(\textbf{x})$ is such that it satisfies the multivariate, higher-order extension to the backward Euler updates given by \eqref{pde}, assuming that the discrete realisations of the sequence are driven by an underlying differential equation; this can be expressed by the following optimisation problem:
\begin{align}\label{regularisation}
& \min \sum_n L(f(\textbf{x}_n), y_n), \quad \text{where} \quad f(\textbf{x}_n) = \sum_{j=1}^{c}w_j\phi_j(\textbf{x}_n)+w_0\nonumber \\
& \text{subject to}: \nonumber \\
& s_{\pi+1} = f(\textbf{x}_n) = \bm{\lambda}^\top\textbf{s}_{\pi-l+1:\pi} + \sum_{k=1}^{\nu}\sum_{i=1}^{d}a_{k,i}\frac{\partial^{(k)}f(\textbf{x}_n)}{\partial x_i^k},
\end{align}
where $L$ is an arbitrary loss function.
This regularisation thus makes the proposed network relatively more robust to noisy observations which affect the choice of RBF centres and widths, causing the vanilla RBF network to underperform in the presence of noise. Consequently, while fitting the data to the differential equation given by \eqref{pde2}, we are implicitly only learning a regularised version of the parameters $w_j$, i.e., regularised by the equality constraint in \eqref{regularisation}.

\subsection{Higher-order derivatives of the RBF}
The fundamental idea in our proposed approach is fitting the data to the backward-Euler-regularised RBF network given by the differential equation in \eqref{pde2}, rather than training the RBF network to approximate $f$ directly as a function of $\textbf{x}$ as in \eqref{rbfn}. The form of \eqref{pde} thus requires us to obtain the expressions for the first and higher-order derivatives of the radial basis function network.

Deriving from \eqref{rbfn3}, the first-order partial derivative of $f$ is given by:
\begin{equation}
    \frac{\partial f(\textbf{x})}{\partial x_i}  = \sum_{j=1}^{c}w_j\frac{\partial \phi_j(\textbf{x})}{\partial x_i},
\end{equation}
where
\begin{equation}\label{phi_first}
    \frac{\partial \phi_j(\textbf{x})}{\partial x_i} = -2\beta_j(x_i-\mu_{j,i}) e^{-\beta_j\|\textbf{x}-\bm{\mu}_j\|^2}
\end{equation}

Accordingly, the second-order partial derivative of $f$ (ignoring mixed derivatives) is given by:
\begin{equation}
    \frac{\partial^2 f(\textbf{x})}{\partial x_i^2}  = \sum_{j=1}^{c}w_j\frac{\partial^2 \phi_j(\textbf{x})}{\partial x_i^2},
\end{equation}
where
\begin{equation}
    \frac{\partial^2 \phi_j(\textbf{x})}{\partial x_i^2} = 2\beta_j \big[2\beta_j(x_i-\mu_{j,i})^2 -1\big]e^{-\beta_j\|\textbf{x}-\bm{\mu}_j\|^2}.
\end{equation}
This generalises to a higher-order $\nu$ as follows:
\begin{equation}
    \frac{\partial^{(\nu)} z(\textbf{x})}{\partial x_i^{\nu}}  = \sum_{j=1}^{c}w_j\frac{\partial^{(\nu)} \phi_j(\textbf{x})}{\partial x_i^{\nu}},
\end{equation}
Similar results have been obtained for the Gaussian RBF \cite{mai2003approximation} and the multiquadric radial basis functions \cite{mai2001numerical}.

Due to the form of the Gaussian radial basis function in \eqref{rbf}, $\phi_j(\textbf{x})$ is infinitely differentiable. In the following, we derive the formula for finding the $k$th derivative of $\phi_j(\textbf{x})$ given by \eqref{rbf}. First, we consider the generalised Leibniz rule for finding the $k$th derivative of the product $uv$ given as:
\begin{equation}
    (uv)^{(k)}=\sum_{m=0}^k{k\choose m}u^{(k-m)}v^{(m)}.
\end{equation}
If we start from the first-order partial derivative of $\phi_j(\textbf{x})$ in \eqref{phi_first}, we can express this as a product $uv$, where:
\begin{equation}
    u = -2\beta_j(x_i-\mu_{j,i}),
\end{equation}
and
\begin{equation}
    v = e^{-\beta_j\|\textbf{x}-\bm{\mu}_j\|^2} = \phi_j(\textbf{x}).
\end{equation}

From Leibniz rule, we can then derive the $k$th order derivative of the RBF as:
\begin{equation}
    \frac{\partial^{(k)}\phi_j(\textbf{x})}{\partial x_i^k}= \begin{cases}
    \phi_j(\textbf{x}), & k = 0\\
    \sum_{m=0}^{k-1} {k-1\choose m} u^{(k-m-1)} \frac{\partial^{(m)}\phi_j(\textbf{x})}{\partial x_i^{(m)}}, & k\geq1
    \end{cases},
\end{equation}
which computes the $k$th partial derivative recursively.

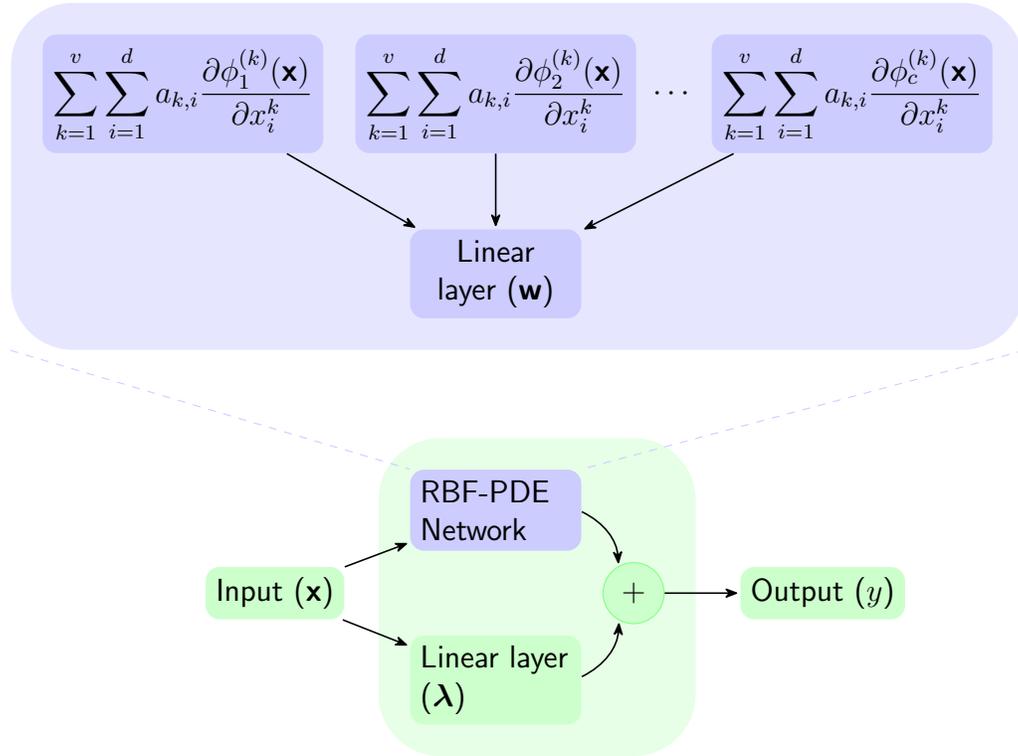
\begin{figure}[h]
\centering
\begin{sffamily}
  \begin{tikzpicture}
    [neuron/.style={rectangle, rounded corners=1ex, fill=blue!20, node distance=1em, on grid=false},
      dataflow/.style={neuron, fill=green!20},
      bend angle=30,
      pre/.style={<-,shorten <=1pt,>={Stealth[round]},semithick}, post/.style={->,shorten >=1pt,>={Stealth[round]},semithick}]

    \node [neuron] (firstneuron) {$\displaystyle \sum_{k=1}^v \sum_{i=1}^d a_{k,i} \frac{\partial \phi_1^{(k)} (\textbf{x})}{\partial x_i^k } $};

    \node [neuron] (secondneuron) [right=of firstneuron] {$\displaystyle \sum_{k=1}^v \sum_{i=1}^d a_{k,i} \frac{\partial \phi_2^{(k)} (\textbf{x})}{\partial x_i^k } $};

    \node [node distance=2pt] (dots) [right=of secondneuron] {$\cdots$};

    \node [neuron, node distance=2pt] (lastneuron) [right=of dots] {$\displaystyle \sum_{k=1}^v \sum_{i=1}^d a_{k,i} \frac{\partial \phi_c^{(k)} (\textbf{x})}{\partial x_i^k } $};

    \node [neuron, node distance=1cm, text width=2cm, align=center] (sum) [below=of secondneuron] {Linear layer ($\textbf{w}$)}
    edge [pre]     (firstneuron)
    edge [pre]     (secondneuron)
    edge [pre]     (lastneuron);

    \begin{scope}[on background layer]
      \node [rectangle, rounded corners=2em, inner sep=1em, fill=blue!10, fit=(firstneuron) (lastneuron) (sum)] (networkbox) {};
    \end{scope}

    \node [neuron, node distance=2cm, text width=2cm] (network) [below=of sum] {RBF-PDE Network};

    \node [minimum width=2cm, node distance=1em] (dummy) [below=of network] {};

    \node [dataflow, text width=2cm] (lambda) [below=of dummy] {Linear layer ($\bm{\lambda}$)};

    \node [dataflow, node distance=1cm] (input) [left=of dummy] {Input ($\textbf{x}$)}
    edge [post]     (network)
    edge [post]     (lambda)
    ;

    \node [circle, draw=green!50, fill=green!20, node distance=1em] (adder) [right=of dummy] {+}
    edge [pre, bend right] (network)
    edge [pre, bend left] (lambda);

    \begin{scope}[on background layer]
      \node [rectangle, rounded corners=2em, inner sep=1em, fill=green!10, fit=(network) (adder) (lambda)] {};
    \end{scope}

    \node [dataflow, node distance=1cm] (output) [right=of adder] {Output ($y$)}
    edge [pre] (adder);
    ;

    \draw [dashed, draw=blue!20] (networkbox.south west) -- (network.north west);
    \draw [dashed, draw=blue!20] (networkbox.south east) -- (network.north east);

  \end{tikzpicture}
\end{sffamily}
\caption{Proposed differential RBF network architecture}
\label{proposed_architecture}
\end{figure}

If we now substitute the radial basis function network $f$ and its higher-order partial derivatives into the PDE in \eqref{pde}, we obtain:
\begin{equation}\label{rbfpde}
    f = \bm{\lambda}^\top\textbf{s}_{\pi-l+1:\pi} + \sum_{j=1}^{c} w_j \sum_{k=1}^{\nu}\sum_{i=1}^{d} a_{k,i}\frac{\partial^{(k)} \phi_j(\textbf{x})}{\partial x_i^k},
\end{equation}
whose network architecture is given in Fig. \ref{proposed_architecture}

In matrix form, \eqref{rbfpde} is equivalent to:
\begin{equation}\label{system}
    f = \bm{\lambda}^\top\textbf{s}_{\pi-l+1:\pi} + \textbf{w}^{\top}\bm{\Theta}(\textbf{x})\textbf{a},
\end{equation}
where,
\begin{equation}\label{w_vec}
    \textbf{w} = [w_1, ..., w_c]^\top,
\end{equation}
\begin{equation}\label{a_vec}
    \textbf{a} = [a_{1,1}, ..., a_{1,d}, ..., a_{\nu,1}, ..., a_{\nu, d}]^\top,
\end{equation}
and
\begin{equation}
    \bm{\Theta}(\textbf{x}) = 
    \begin{bmatrix}
    \frac{\partial^{(1)} \phi_1(\textbf{x})}{\partial x_1} & \cdots & \frac{\partial^{(1)}\phi_1(\textbf{x})}{\partial x_d} & \cdots & \frac{\partial^{(\nu)} \phi_1(\textbf{x})}{\partial x_1^{\nu}} & \cdots & \frac{\partial^{(\nu)} \phi_1(\textbf{x})}{\partial x_d^\nu}\\
    \vdots & \ddots & \vdots & \ddots & \vdots & \ddots & \vdots \\
    \frac{\partial^{(1)} \phi_c(\textbf{x})}{\partial x_1} & \cdots & \frac{\partial^{(1)}\phi_c(\textbf{x})}{\partial x_d} & \cdots & \frac{\partial^{(\nu)} \phi_c(\textbf{x})}{\partial x_1^{\nu}} & \cdots & \frac{\partial^{(\nu)} \phi_c(\textbf{x})}{\partial x_d^\nu}
    \end{bmatrix}
\end{equation}

Using \eqref{system}, we may then optimise $\textbf{w}, \bm{\lambda}$ and $\textbf{a}$ jointly on the training data for arbitrary objective functions. For example, for the mean-squared error $\epsilon$ given as:
\begin{equation}
    \epsilon = \frac{1}{N}\sum_{n=1}^{N} \big(y_n - \textbf{w}^\top \bm{\phi}(\textbf{x}_n) - w_0 - \textbf{w}^{\top}\bm{\Theta}(\textbf{x}_n)\textbf{a}\big)^2,
\end{equation}
we can optimise the network weights $\textbf{w}, \bm{\lambda}$ and $\textbf{a}$ to an arbitrary accuracy with a given choice of an optimiser such as stochastic gradient descent (SGD).

\section{Experimental Validation}
In this section, we test the proposed differential RBF network on the logistic map chaotic timeseries \cite{maathuis2017predicting,farmer1987predicting}, as well as on $30$ different timeseries from the M5 competition \cite{makridakis2020m5}; the reasons for the choice of these datasets are given in Sections \ref{log_map_exp} and \ref{m5_exp} respectively.

\subsection{Logistic map}\label{log_map_exp}
We consider the logistic map given as:
\begin{equation}\label{logistic_map}
s_{\pi+1} = 4s_{\pi}[1-s_{\pi}],
\end{equation}
and we generate $1000$ observations in the forward pass as our timeseries using an initial value of $0.1$; the first $200$ observations are shown in Figure \ref{log_map}:
\begin{figure}
    \centering
    \includegraphics[width=0.8\textwidth]{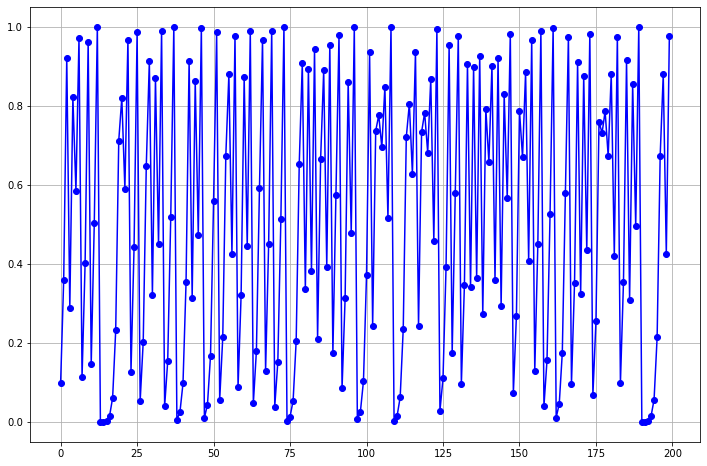}
    \caption{Logistic map chaotic timeseries}
    \label{log_map}
\end{figure}
We split the timeseries into training and testing sets, with the last $100$ observations as the test set. We have selected the logistic map because, although it is an archetypal chaotic series, it arises from simple polynomial mappings that can be easily modelled by the RBF network. Furthermore, to simulate observational noise, we add some Gaussian noise with variance $\omega \in \{0, 0.02, 0.04, 0.08, 0.12\}$ to the timeseries (whose original chaotic behaviour is not due to noise) in order to demonstrate the noise susceptibility of the RBF network. We have selected $\omega$ so that it does not exceed the variance of the original timeseries which is $0.12$, in order not to corrupt the information contained in the series.  We perform mean-normalisation, and then reshape the training data into input-output pairs using a lookback window of $l\in\{1,2,4,8,16\}$ with which we train the unnormalised and normalised RBF networks and the proposed differential RBF network. Although $l=1$ suffices for the logistic map because it depends only its previous input, as shown in \eqref{logistic_map}, we experiment with exponentially increasing $l$ in order to demonstrate the replication of noisy observations as inputs to the RBF network. The number of RBF centres used is $c= \max(5, 2l)$, which is what showed the best performance after experimenting with different configurations. On the test set, we perform one-step prediction and report the mean absolute error (MAE). We defer multi-step prediction to Section \ref{m5_exp}.

For all the RBF networks (i.e., unnormalised, normalised and differential variants), we determine the examplars $\bm{\mu}_j$ via K-Means clustering while the RBF parameters $\beta_j$ are defined according to \eqref{rbf_width}. We minimise the least-squares error with lasso regularisation, with the regularisation coefficient determined via cross-validation. Furthermore, the order $\nu$ of the PDE in the differential RBF network is set empirically to $\nu=2$ based on cross-validation results, and we initialise the differential RBF network weights $\textbf{w}, \bm{\lambda}$ and $\textbf{a}$ with the following settings:
\begin{align}
    &\bm{\lambda} = \frac{1}{l}\label{lambda_wts}, \\
    &a_{i,k} = \frac{0.001^k}{k!} \label{a_wts}, \\
    &\textbf{w} = [\tilde{w}_1^*, ..., \tilde{w}_c^*]^\top \label{w_wts},
\end{align}
where ${i,k}$ index the $k$th-order partial derivative of the RBF with respect to the component $x_i$ of the input $\textbf{x}$, and $\tilde{w}_1^*, ..., \tilde{w}_c^*$ are obtained from \eqref{w_opt} with lasso regularisation, given that $\tilde{\textbf{w}}^* = [\tilde{w}_0^*, \tilde{w}_1^*, ..., \tilde{w}_c^*]^\top$. The above choices of weight initialisations for the proposed network are justified as follows:
\begin{enumerate}
    \item By setting $\bm{\lambda}$ as in \eqref{lambda_wts}, equal weights are assigned to the last $l$ values of the sequence.
    \item By setting $a_{i,k}$ as in \eqref{a_wts}, the PDE weights are defined as Taylor expansion coefficients with $0.001$ being the interval between any two realisations of the series.
    \item By setting $\textbf{w}$ as in \eqref{w_wts}, we utilise the RBF weights obtained from the unnormalised RBF network.
\end{enumerate}
We subsequently optimise the network weights in Figure \ref{proposed_architecture} using the Broyden, Fletcher, Goldfarb, and Shanno (BFGS) method \cite{nocedal2006numerical}.

The MAE results from these experiments are given in Tables \ref{log_map_table1} to \ref{log_map_table5}, while the model predictions are given in the appendix in Figures \ref{log_1} to \ref{log_5}.

\begin{table}[tbph!]
    \begin{multicols}{2}
    \centering
    \begin{tabular}{R{0.10}R{0.2}R{0.2}R{0.2}}
\toprule
 $l$ &    RBF-DiffNet &   u-RBFN &         n-RBFN \\
\midrule
   1 &        $0.277$ &  $0.146$ &  $\bf{0.0994}$ \\
   2 &   $\bf{0.133}$ &   $0.16$ &        $0.166$ \\
   4 &  $\bf{0.0879}$ &  $0.277$ &        $0.311$ \\
   8 &  $\bf{0.0717}$ &  $0.223$ &        $0.313$ \\
  16 &  $\bf{0.0542}$ &  $0.236$ &        $0.314$ \\
\bottomrule
\end{tabular}

    \caption{Noise variance $\omega = 0$}
    \label{log_map_table1}
    
    \begin{tabular}{R{0.10}R{0.2}R{0.2}R{0.2}}
\toprule
 $l$ &    RBF-DiffNet &   u-RBFN &        n-RBFN \\
\midrule
   1 &        $0.286$ &  $0.152$ &  $\bf{0.103}$ \\
   2 &   $\bf{0.124}$ &  $0.162$ &       $0.167$ \\
   4 &   $\bf{0.082}$ &  $0.283$ &       $0.312$ \\
   8 &  $\bf{0.0648}$ &  $0.199$ &       $0.308$ \\
  16 &  $\bf{0.0517}$ &  $0.221$ &       $0.315$ \\
\bottomrule
\end{tabular}

    \caption{Noise variance $\omega = 0.02$}
    \label{log_map_table2}
    \end{multicols}
\end{table}

\begin{table}[tbph!]
    \begin{multicols}{2}
    \centering
    \begin{tabular}{R{0.10}R{0.2}R{0.2}R{0.2}}
\toprule
 $l$ &    RBF-DiffNet &   u-RBFN &         n-RBFN \\
\midrule
   1 &        $0.286$ &   $0.14$ &  $\bf{0.0987}$ \\
   2 &   $\bf{0.143}$ &  $0.161$ &        $0.157$ \\
   4 &  $\bf{0.0582}$ &  $0.291$ &        $0.313$ \\
   8 &   $\bf{0.063}$ &  $0.216$ &        $0.315$ \\
  16 &  $\bf{0.0631}$ &  $0.244$ &        $0.314$ \\
\bottomrule
\end{tabular}

    \caption{Noise variance $\omega = 0.04$}
    \label{log_map_table3}
    
    \begin{tabular}{R{0.10}R{0.2}R{0.2}R{0.2}}
\toprule
 $l$ &    RBF-DiffNet &   u-RBFN &         n-RBFN \\
\midrule
   1 &          $0.3$ &  $0.109$ &  $\bf{0.0916}$ \\
   2 &        $0.151$ &  $0.139$ &   $\bf{0.132}$ \\
   4 &  $\bf{0.0659}$ &  $0.301$ &        $0.309$ \\
   8 &   $\bf{0.107}$ &  $0.248$ &        $0.312$ \\
  16 &    $\bf{0.11}$ &  $0.293$ &        $0.315$ \\
\bottomrule
\end{tabular}

    \caption{Noise variance $\omega = 0.08$}
    \label{log_map_table4}
    \end{multicols}
\end{table}

\begin{table}[tbph!]
    \centering
    
    \begin{tabular}{R{0.10}R{0.2}R{0.2}R{0.2}}
\toprule
 $l$ &    RBF-DiffNet &   u-RBFN &        n-RBFN \\
\midrule
   1 &        $0.295$ &  $0.138$ &  $\bf{0.106}$ \\
   2 &         $0.18$ &   $0.15$ &  $\bf{0.144}$ \\
   4 &  $\bf{0.0852}$ &  $0.299$ &        $0.31$ \\
   8 &   $\bf{0.141}$ &  $0.275$ &       $0.309$ \\
  16 &   $\bf{0.162}$ &  $0.293$ &       $0.314$ \\
\bottomrule
\end{tabular}

    \caption{Noise variance $\omega = 0.12$}
    \label{log_map_table5}
    \caption*{Mean absolute error (MAE) performance on the logistic map at different noise levels. Best values are in boldface. $l$ is the lookback window; RBF-DiffNet is the proposed differential RBF network; u-RBFN and n-RBFN are the unnormalised and normalised RBF networks respectively.}
\end{table}

\subsection{M5 dataset}\label{m5_exp}
The M5 competition is the latest in a series of M-forecasting competitions organised by the University of Nicosia that has drawn the interest of top forecasting practitioners including Google \cite{fry2019m4} and Uber \cite{smyl2020hybrid}. The M5 competition comprises $42840$ individual timeseries data made available by Walmart. This dataset consists of unit sales of products across the United States at different levels of aggregation. This dataset has been chosen because, like many real-world datasets, it exhibits a non-deterministic (noisy) behaviour as shown in Figures \ref{m5_1} to \ref{m5_5}. Furthermore, by not including exogenous inputs such as prices that can affect product sales, we impose further \textit{deterministic noise} \cite{abu2012learning} on the series. Thus, the dataset is able to sufficiently illustrate the susceptibility of the baseline RBF network to noisy inputs in sequential data.

Due to computational constraints, we work at the level $8$ aggregation that consists of $30$ different timeseries data of unit sales of products aggregated for each of $10$ stores (across three US states) and $3$ product categories (i.e., foods, household and hobbies). Each of the $30$ different timeseries has $1941$ daily observations (spanning the period between 29 January 2011 and 19 June 2016), with the last $28$ days set for validation.

Different benchmark algorithms (categorised under statistical, machine learning and combinations of both) have been provided by the M5 competition organisers, and in this section, we limit our comparison to one statistical benchmark: autoregressive integrated moving average (ARIMA), and one machine learning benchmark: the multi-layer perceptron (MLP). The benchmark MLP architecture specified in the M5 competition is given as follows \cite{makridakis2020m5}:
\begin{enumerate}
    \item Number of input layer nodes = $14$, corresponding to the last $14$ days of sales data
    \item Number of hidden layer nodes = $28$,
    \item Number of output nodes = $1$
    \item Hidden layer activation function: logistic; output layer activation function: linear,
    \item Number of iterations = $500$,
    \item Ensemble size = $10$, i.e, $10$ different MLPs are trained and their predictions aggregated in terms of the median operator to account for poor weight initialisations.
\end{enumerate}
Using the above network parameters, we also implement an ensemble of $10$ recurrent networks with LSTM blocks which use tanh activation for further comparison.

To make the RBF networks comparable to the MLP architecture, we set $d$ (the RBF network input size) to $14$ and $c$ (the number of RBF nodes) to $28$. The same settings of the RBF networks used in Section \ref{log_map_exp} are used here; the order $\nu$ of the PDE in the differential RBF network is however set to empirically to $\nu=1$, based on cross-validation results.

For each timeseries, we set the last $28$ observations as the test set, and the remaining as the training set. Since preprocessing for timeseries forecasting typically involves stationarising the series to make the series easier to predict \cite{gheyas2009neural,pal2017practical}, we first perform a stationarity test on the training set using the augmented Dickey-Fuller test \cite{mushtaq2011augmented}; if the timeseries does not exhibit stationarity according to the test's null hypothesis, we proceed to difference the timeseries incrementally until it shows stationarity, up to a differencing order of $10$. We then perform mean-normalisation and then reshape the data into input-output pairs using a lookback window of $l=14$.

On the test set, we perform reverse-differencing and normalisation operations to recover the original series and range, and we perform multi-step prediction over the forecast horizon of $28$ days. We report the root mean squared scaled error (RMSSE), which is the error metric used in the M5 competition. The RMSSE is defined as:
\begin{equation}
\text{RMSSE} = \frac{(n_{train}- 1)}{n_{test}}\frac{\sum_{\pi=n_{train}+1}^{n_{train}+n_{test}}(s_{\pi}-\hat{s}_{\pi})^2}{\sum_{\pi=2}^{n_{train}}(s_{\pi}-s_{\pi-1})^2},
\end{equation}
where $\hat{s}_{\pi}$ is a model's estimate of the timeseries at time $t_\pi$, $n_{train}=1913$ is the length of training series, and $n_{test}=28$ is the length of the forecast horizon.

We do not investigate cross-learning, where information from other timeseries are used in training a single global model to predict each timeseries \cite{makridakis2020m4}; this is out of the scope of this paper.


\subsection{Results and discussion}

\subsubsection{Logistic map data}\label{log_discussions}
From the results in Tables \ref{log_map_table1} to \ref{log_map_table5}, there is an increase in the MAE values for the unnormalised RBF network (u-RBFN) as the lookback window $l$ increases. This is especially observed for high noise levels $\omega$, since the noisy observations gets replicated many times in the reshaped data that is input to the RBF network, illustrating the susceptibility of the unnormalised RBF network to noisy observations.

The normalised RBF network (n-RBFN), on the other hand, shows robustness in its predictions across different noise levels, with the MAE values remaining relatively constant. However, as the lookback window length $l$ increases, its performance also deteriorates for each of the noise levels. This performance degradation is mainly due to the side effects of normalisation \cite{shorten1996side,shorten1994normalising} whereby the maxima of the RBF shifts when the RBFs have unequally spaced centres or different widths. As Shorten \cite{shorten1994normalising} notes, this side effect becomes more pronounced with increase in the input dimension. This causes points far away from the RBF centre to have high functional values, and thus, rather than a few RBFs getting activated, many more RBFs tend to contribute uniformly to the output for any given input. This is manifested in the n-RBFN output approaching the mean of the timeseries for large values of $l$ as shown in Figures \ref{log_1} to \ref{log_5}.

The differential RBF network (RBF-DiffNet), however, shows robustness across different noise variance $\omega$ and different values of $l$, significantly outperforming the benchmark RBF networks. In particular, for low to medium noise levels, RBF-DiffNet is able to utilise more information in the lagged values of the series as $l$ increases to improve the prediction accuracy, without any noise enhancement or side effects of normalisation, unlike for u-RBFN and n-RBFN. While the MAE performance can be seen to degrade as the noise variance increases, this is to be expected as the information in the timeseries increasingly gets corrupted. However, RBF-DiffNet noticeably underperforms for $l=1$ across all noise levels; this is likely due to the fact that the regularisation due to the backward-Euler updates causes the RBF-DiffNet to be overly smooth resulting in underfitting, even though the reshaped data input to the RBF network may not contain many noisy observations for $l=1$. Thus, RBF-DiffNet is most suited to timeseries forecasting tasks for which the series to be predicted is noisy and influenced by more than one previous realisations of the series: an example of this is demand forecasting of grocery items, where future sales may be autocorrelated with historical sales up to $l=14$ days prior; other examples include weather forecasting or sentence completion in natural language processing.

In terms of computational performance, u-RBFN and n-RNFN having closed-form solutions given in \eqref{w_opt} outperform the proposed RBF-DiffNet which requires an iterative optimisation routine. For a large number of RBF nodes, the solution given by \eqref{w_opt} may be computationally intractable and u-RBFN and n-RBFN may require iterative optimisation techniques such as SGD. Since u-RBFN and n-RBFN each has $c+1$ optimising variables as compared to $c+d\nu+l+1$ for the differential RBF network, the existing RBF networks may still have a computational edge over RBF-DiffNet in an iterative optimisation technique.

\subsubsection{M5 datasets}\label{m5_discussions}
Although the different models are able to maintain relatively constant prediction accuracy across the forecast horizon as shown in Figures \ref{m5_1} to \ref{m5_5} due to the seasonalities in the timeseries, the results in Table \ref{m5_endog} show that the differential RBF network (RBF-DiffNet) achieves superior predictive accuracy over the unnormalised RBF network (u-RBFN) and normalised RBF network (n-RBFN). Over the $30$ different timeseries, RBF-DiffNet achieves a $26\%$ reduction over u-RBFN and an $18\%$ reduction over n-RBFN in terms of the mean RMSSE. This performance improvement is shown to be statistically significant at the $0.95$ confidence level, as seen from the p-values in Table \ref{wilcoxon}.

The RBF-DiffNet is also shown to yield a $4\%$ reduction over ARIMA (significant at the $0.95$ confidence level), $0.2\%$ over the MLP (statistically insignificant), and $2.2\%$ over the LSTM (statistically insignificant) in terms of the RMSSE.

The reason for the performance gain of RBF-DiffNet over u-RBFN and n-RBFN is the relative robustness of RBF-DiffNet to noise as described in Section \ref{log_discussions} due to its backward-Euler regularisation, while the performance gain of RBF-DiffNet over ARIMA is because the RBF network generally learns complex non-linear mappings as compared to the simple linear mapping in the ARIMA model.

Despite their comparable performance, it is worth noting that RBF-DiffNet uses only one weight initialisation as given in \eqref{lambda_wts}, \eqref{a_wts} and \eqref{w_wts}, whereas the LSTM and MLP ensembles respectively use $10$ different weight initialisations and the median operator to mitigate the effects of poor weight initialisations. The single weight initialisation is possible in RBF-DiffNet because its network weights have more intuitive meanings which allow for an informed choice of initialisation in such a way that multiple random restarts yield only minor performance improvements. Consequently, for the comparable performance given in Table \ref{m5_endog}, the RBF network requires at least $16$ times less computational time than the LSTM ensemble, as shown in Table \ref{compute_time}. Moreover, RBF-DiffNet has the added advantage of being interpretable in terms of the influence of different input features on the output, since the hidden layer performs a fuzzy nearest-neighbour association of a given input to the RBF centres.

\begin{table}[tbph!]
    \centering
    \begin{tabular}{R{0.10} R{0.11} R{0.11} R{0.11} R{0.11} R{0.11} R{0.11} }
\toprule
Series ID &       RBF-DiffNet &            u-RBFN &            n-RBFN &         ARIMA &           MLP &          LSTM \\
\midrule
        1 &             $0.5$ &  $\mathit{0.461}$ &            $0.53$ &       $0.513$ &  $\bf{0.433}$ &        $0.46$ \\
        2 &  $\mathit{0.693}$ &            $0.78$ &           $0.701$ &  $\bf{0.686}$ &        $0.76$ &       $0.749$ \\
        3 &      $\bf{0.422}$ &           $0.621$ &           $0.532$ &       $0.512$ &       $0.467$ &       $0.489$ \\
        4 &           $0.761$ &            $1.85$ &  $\mathit{0.695}$ &       $0.867$ &       $0.801$ &  $\bf{0.683}$ \\
        5 &  $\mathit{0.702}$ &           $0.992$ &           $0.972$ &       $0.884$ &       $0.833$ &  $\bf{0.581}$ \\
        6 &       $\bf{0.54}$ &           $0.745$ &           $0.564$ &       $0.749$ &       $0.664$ &       $0.598$ \\
        7 &  $\mathit{0.625}$ &           $0.662$ &            $1.35$ &       $0.492$ &  $\bf{0.438}$ &       $0.653$ \\
        8 &  $\mathit{0.587}$ &            $1.47$ &           $0.786$ &       $0.582$ &  $\bf{0.552}$ &       $0.554$ \\
        9 &           $0.609$ &            $1.33$ &  $\mathit{0.538}$ &       $0.556$ &  $\bf{0.528}$ &       $0.616$ \\
       10 &  $\mathit{0.751}$ &           $0.771$ &            $1.45$ &       $0.691$ &       $0.743$ &  $\bf{0.663}$ \\
       11 &  $\mathit{0.945}$ &            $1.14$ &            $1.09$ &       $0.978$ &       $0.935$ &  $\bf{0.928}$ \\
       12 &  $\mathit{0.895}$ &            $2.25$ &            $1.76$ &  $\bf{0.875}$ &        $1.12$ &        $1.16$ \\
       13 &  $\mathit{0.859}$ &           $0.921$ &           $0.936$ &       $0.862$ &  $\bf{0.697}$ &           $1$ \\
       14 &  $\mathit{0.746}$ &           $0.945$ &           $0.775$ &       $0.923$ &       $0.759$ &  $\bf{0.725}$ \\
       15 &           $0.847$ &  $\mathit{0.836}$ &            $1.02$ &   $\bf{0.83}$ &       $0.837$ &        $1.03$ \\
       16 &  $\mathit{0.724}$ &             $0.9$ &            $1.38$ &        $0.55$ &  $\bf{0.521}$ &       $0.571$ \\
       17 &  $\mathit{0.663}$ &           $0.796$ &           $0.704$ &       $0.667$ &       $0.686$ &  $\bf{0.637}$ \\
       18 &      $\bf{0.526}$ &           $0.967$ &           $0.643$ &       $0.536$ &       $0.578$ &       $0.659$ \\
       19 &            $1.55$ &       $\bf{0.85}$ &            $1.37$ &        $1.11$ &        $1.38$ &        $1.12$ \\
       20 &      $\bf{0.783}$ &            $1.48$ &           $0.977$ &       $0.916$ &       $0.877$ &       $0.817$ \\
       21 &           $0.858$ &            $1.07$ &      $\bf{0.661}$ &       $0.809$ &        $0.75$ &        $1.33$ \\
       22 &      $\bf{0.627}$ &           $0.778$ &           $0.719$ &       $0.753$ &       $0.704$ &       $0.634$ \\
       23 &      $\bf{0.466}$ &           $0.489$ &           $0.472$ &       $0.522$ &       $0.648$ &       $0.485$ \\
       24 &           $0.637$ &      $\bf{0.625}$ &           $0.674$ &       $0.687$ &        $0.78$ &       $0.788$ \\
       25 &            $1.45$ &       $\bf{1.42}$ &            $1.52$ &        $1.87$ &        $1.46$ &        $1.44$ \\
       26 &       $\bf{0.68}$ &           $0.841$ &           $0.973$ &       $0.741$ &       $0.731$ &        $0.73$ \\
       27 &   $\mathit{1.02}$ &            $1.95$ &            $1.54$ &        $1.12$ &       $0.988$ &  $\bf{0.975}$ \\
       28 &           $0.816$ &           $0.971$ &  $\mathit{0.793}$ &        $1.03$ &  $\bf{0.705}$ &       $0.761$ \\
       29 &           $0.808$ &  $\mathit{0.797}$ &           $0.858$ &       $0.821$ &  $\bf{0.778}$ &        $0.81$ \\
       30 &  $\mathit{0.676}$ &            $1.04$ &           $0.849$ &       $0.697$ &        $0.67$ &  $\bf{0.639}$ \\
     Mean &      $\bf{0.759}$ &            $1.02$ &           $0.927$ &       $0.794$ &        $0.76$ &       $0.776$ \\
   Median &  $\mathit{0.713}$ &           $0.911$ &           $0.821$ &       $0.751$ &       $0.737$ &  $\bf{0.704}$ \\
\bottomrule
\end{tabular}

    \caption{Root mean squared scaled error (RMSSE) results on the M5 dataset (level 8 aggregation). RBF-DiffNet is the proposed differential RBF network; u-RBFN and n-RBFN are the unnormalised and normalised RBF networks respectively. Best values across all algorithms are in boldface. Best values across the three RBF networks (RBF-DiffNet, u-RBFN, n-RBFN) are in italics.}
    \label{m5_endog}
\end{table}
\
\setlength{\arrayrulewidth}{0.3mm}
\begin{table}[tbph!]
    \centering
    \begin{tabular}{cc}
        \toprule
        Algorithm &  p-value\\
        \midrule
        u-RBFN & $\num{3.473e-5}$\\
        n-RBFN & $2.098\times 10^{-4}$\\
        ARIMA & $2.619\times 10^{-2}$\\
        MLP & $4.300\times 10^{-1}$\\
        LSTM & $2.783\times 10^{-1}$\\
        \bottomrule
    \end{tabular}
    \caption{p-values from Wilcoxon signed rank test for differences in median between proposed differential RBF network (RBF-DiffNet) and other algorithms. u-RBFN and n-RBFN are the unnormalised and normalised RBF networks respectively.}
    \label{wilcoxon}
\end{table}

\begin{table}[tbph!]
    \centering
    \begin{tabular}{cc}
        \toprule
        Algorithm &  Training time (seconds)\\
        \midrule
        RBF-DiffNet & $38.97\pm 0.39$\\
        u-RBFN & $1.27\pm 0.19$\\
        n-RBFN & $1.20\pm 0.16$\\
        ARIMA & $33.34\pm 10.74$\\
        MLP & $307.23\pm 4.30$\\
        LSTM & $633.61\pm 6.84$\\
        \bottomrule
    \end{tabular}
    \caption{Model training time on on the M5 datasets (level 8 aggregation) given in the form: mean $\pm$ s.d. obtained on a 32GB-RAM Tesla P100-PCIE 128GB GPU processor with 8 Intel E5-2650 CPU cores @2.2GHz. RBF-DiffNet is the proposed differential RBF network; u-RBFN and n-RBFN are the unnormalised and normalised RBF networks respectively.}
    \label{compute_time}
\end{table}

\section{Conclusion}
The radial basis function (RBF) network has good function approximation properties, fast training time and is easily interpretable in terms of the influence of different input features on the output. However, it is especially sensitive to noisy inputs due to the usually unsupervised step in selecting the RBF centres and widths. This paper extends the RBF network for use on noisy sequential data. For this purpose, we have introduced a novel differential RBF network (RBF-DiffNet) whose hidden layer blocks are higher-order constant-coefficient partial differential equations in terms of the RBF. RBF-DiffNet exhibits robustness to noisy perturbations to the input sequence by regularising the network following backward-Euler updates, assuming the sequence data originates from an underlying differential equation. Notably, RBF-DiffNet differs from the neural ordinary differential equation (ODE-Net) in that RBF-DiffNet parameterises the solution to the underlying differential equation with an RBF network and directly evaluates the derivatives of the network based on the backward-Euler discretisation, while the ODE-Net parameterises the derivatives of the hidden state of a residual network with another neural network.

The proposed differential RBF network is experimentally validated on the logistic map and $30$ other real-world and noisy timeseries data from the M5 competition (level 8 aggregation). Experimental results show RBF-DiffNet significantly outperforms the normalised and unnormalised RBF networks at different noise levels on the logistic map. On the M5 dataset, RBF-DiffNet outperforms the normalised and unnormalised RBF networks by $16\%$ and $28\%$ respectively; furthermore, because its network weights have intuitive meanings, RBF-DiffNet allows for an informed choice of initialisation for the network weights, thereby achieving comparable performance to an ensemble of $10$ LSTM networks (built from $10$ different weight initialisations) at less than one-sixteenth the LSTM computational time.

Our proposed network therefore enables more accurate predictions\textemdash in spite of the presence of observational noise\textemdash in sequence modelling tasks such as timeseries forecasting that leverage the fast training and function approximation properties of the RBF network, where model interpretability is desired.

Nevertheless, the current development of RBF-DiffNet is limited to scalar outputs, and on-going work is investigating its extension to vector outputs. The feasibility of the proposed approach to cross-learning, where information from multiple series is used in training a single global model, is also an area we are currently researching.

\bibliographystyle{elsarticle-num}
\bibliography{references}

\appendix
\section{Logistic map results}
This section contains graphical results from model predictions on the logistic map at different noise levels as described in Section \ref{log_map_exp}.
\begin{figure}[tbph]
    \centering
    \subfloat[$\omega = 0$, $l=1$]{\includegraphics[width=0.5\textwidth]{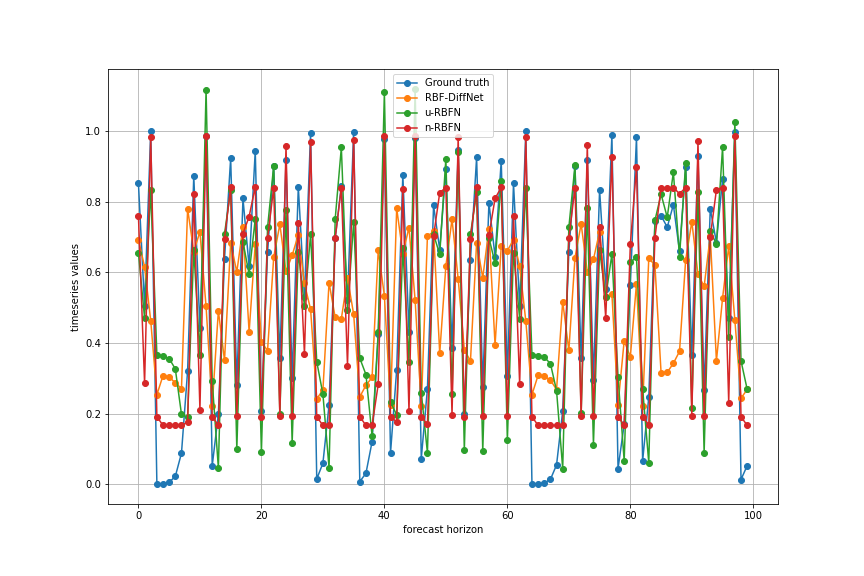}}
    \subfloat[$\omega = 0$, $l=2$]{\includegraphics[width=0.5\textwidth]{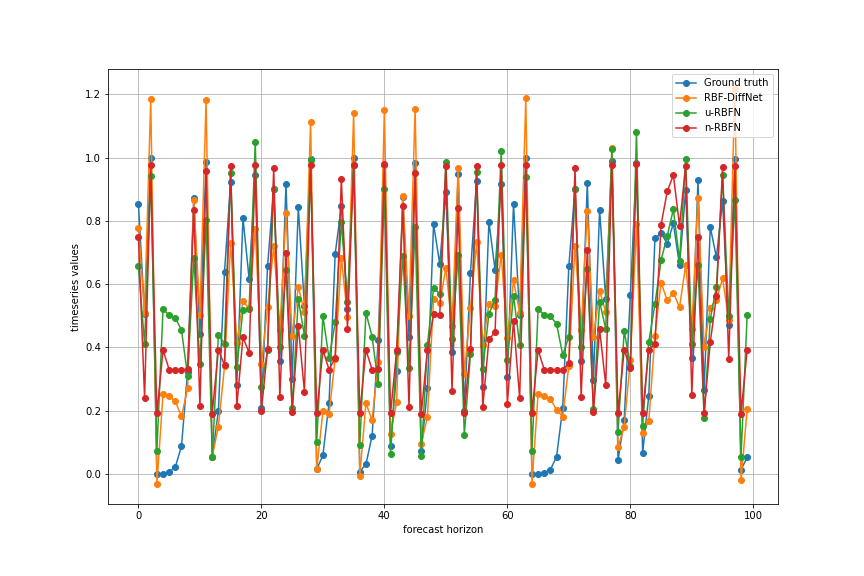}}\\
    \subfloat[$\omega = 0$, $l=4$]{\includegraphics[width=0.5\textwidth]{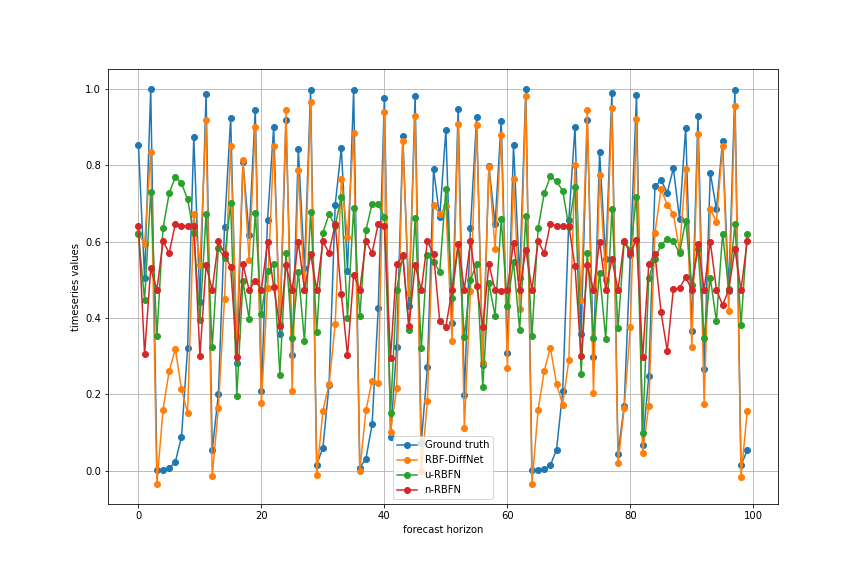}}
    \subfloat[$\omega = 0$, $l=8$]{\includegraphics[width=0.5\textwidth]{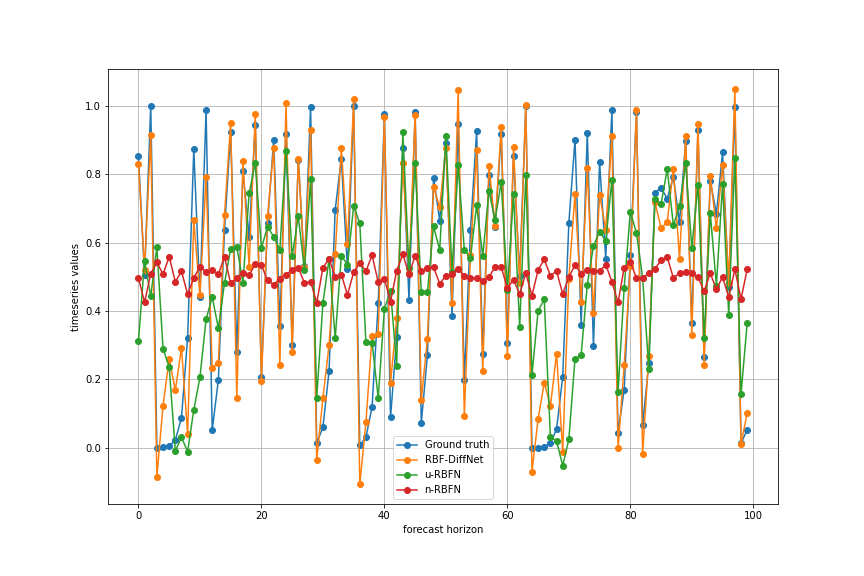}}\\
    \subfloat[$\omega = 0$, $l=16$]{\includegraphics[width=0.5\textwidth]{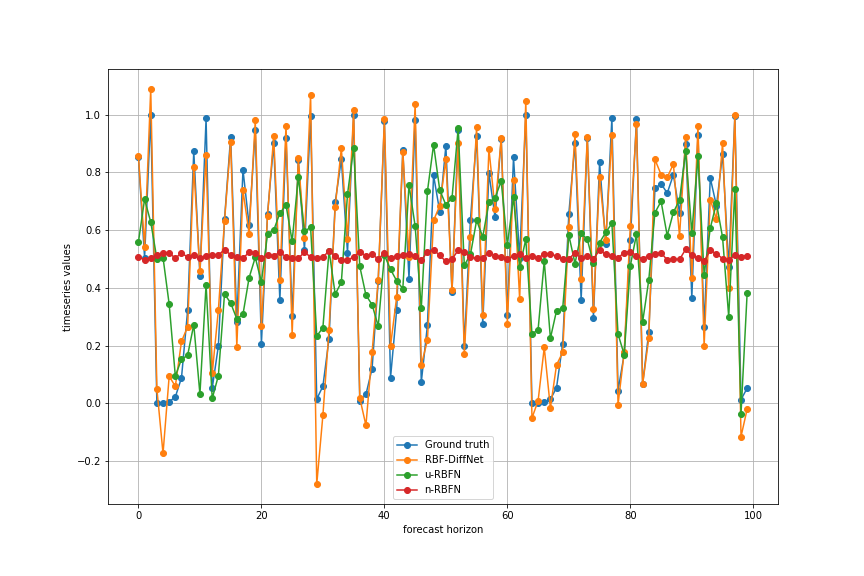}}
    \subfloat[$\omega = 0.02$, $l=1$]{\includegraphics[width=0.5\textwidth]{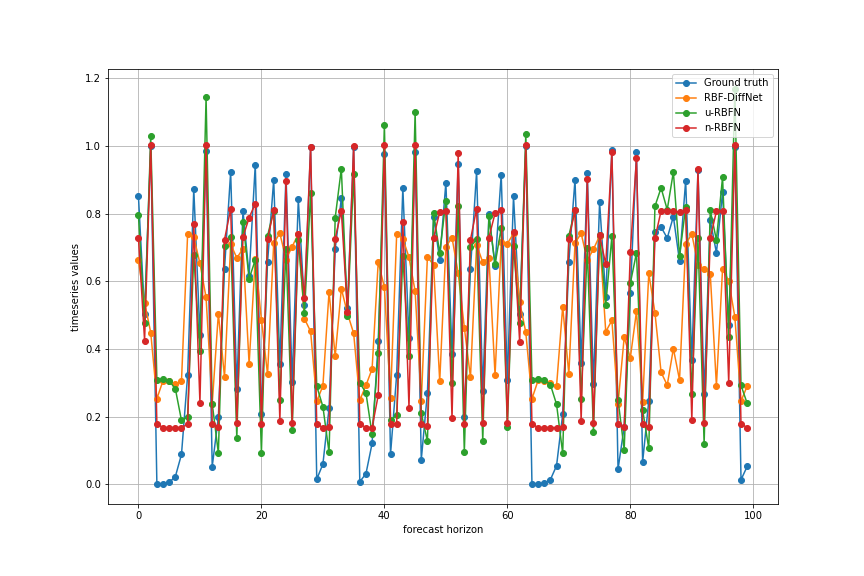}}
    \caption{Model predictions on the logistic map at different noise levels. RBF-DiffNet is the proposed differential RBF network; u-RBFN and n-RBFN are the unnormalised and normalised RBF networks respectively. $\omega$ is the noise variance; $l$ is the lookback window length.}
    \label{log_1}
\end{figure}

\begin{figure}[tbph]
    \centering
    \subfloat[$\omega = 0.02$, $l=2$]{\includegraphics[width=0.5\textwidth]{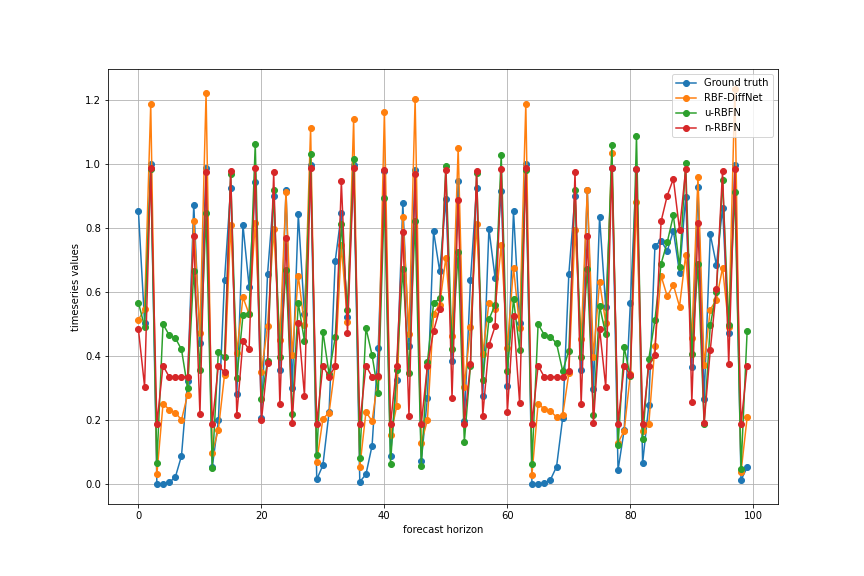}}
    \subfloat[$\omega = 0.02$, $l=4$]{\includegraphics[width=0.5\textwidth]{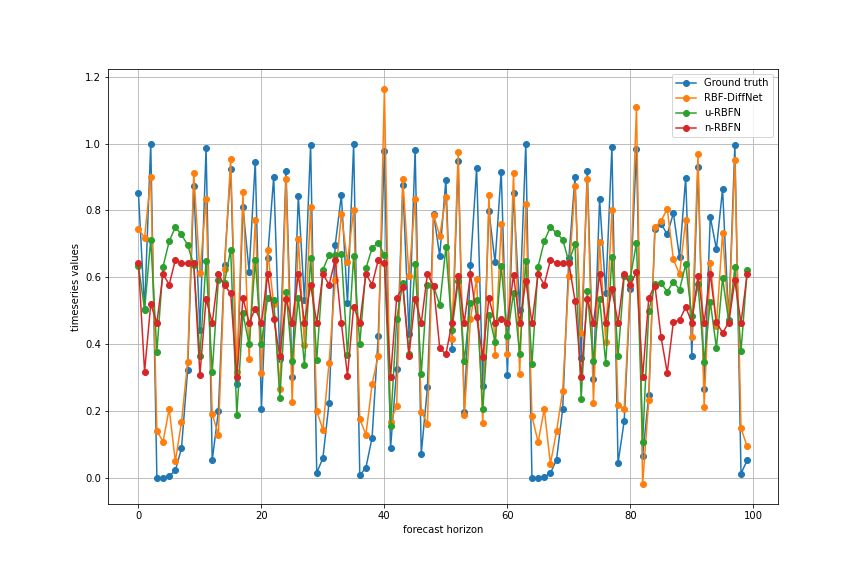}}\\
    \subfloat[$\omega = 0.02$, $l=8$]{\includegraphics[width=0.5\textwidth]{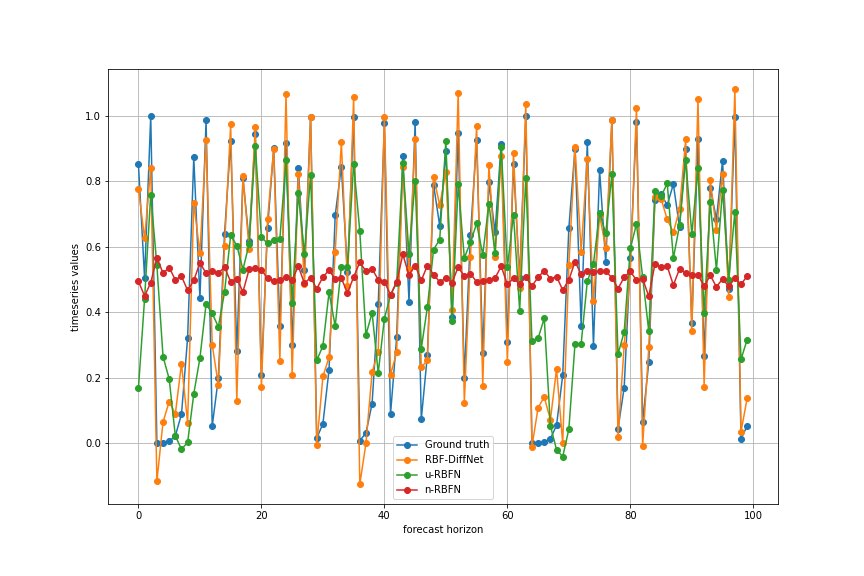}}
    \subfloat[$\omega = 0.02$, $l=16$]{\includegraphics[width=0.5\textwidth]{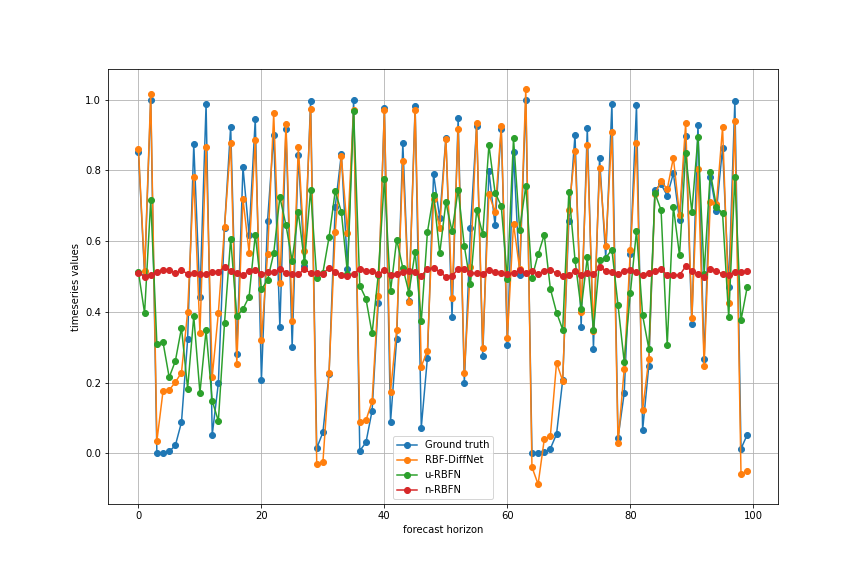}}\\
    \subfloat[$\omega = 0.04$, $l=1$]{\includegraphics[width=0.5\textwidth]{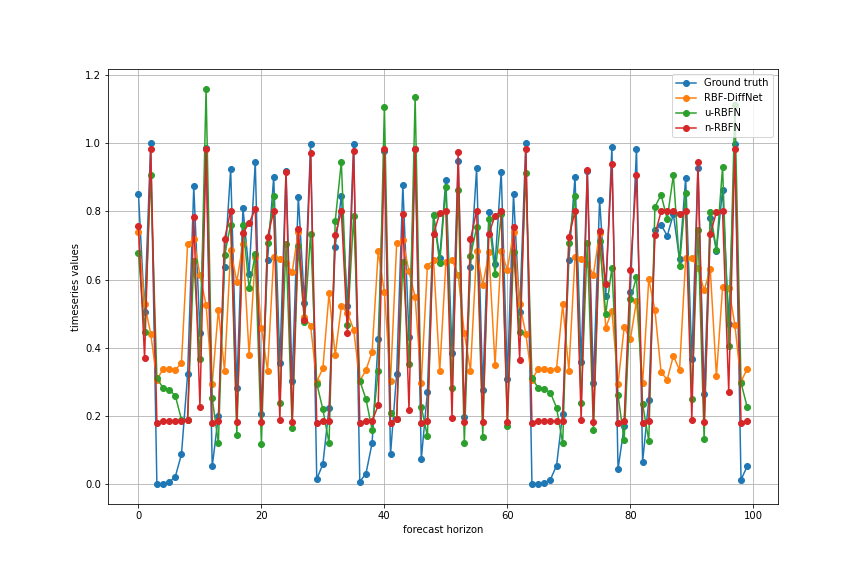}}
    \subfloat[$\omega = 0.04$, $l=2$]{\includegraphics[width=0.5\textwidth]{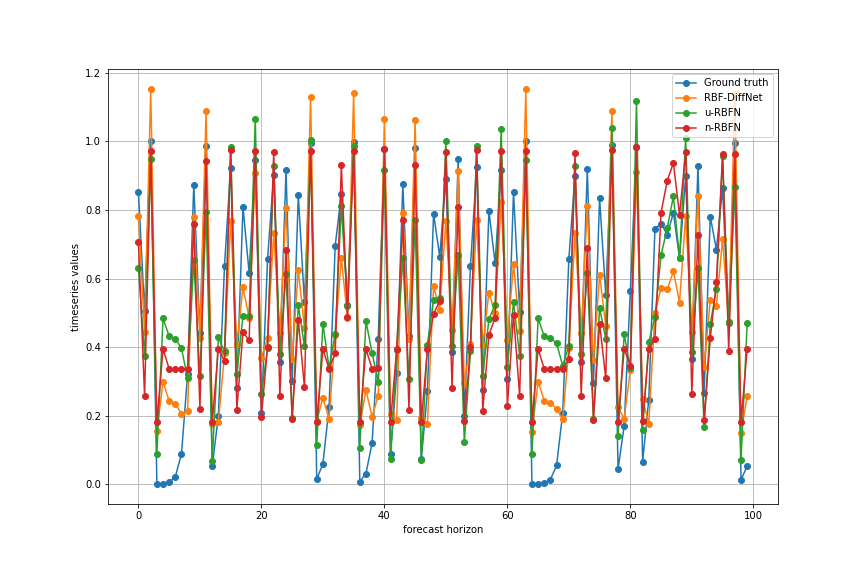}}
    \caption{Model predictions on the logistic map at different noise levels. RBF-DiffNet is the proposed differential RBF network; u-RBFN and n-RBFN are the unnormalised and normalised RBF networks respectively. $\omega$ is the noise variance; $l$ is the lookback window length.}
    \label{log_2}
\end{figure}

\begin{figure}[tbph]
    \centering
    \subfloat[$\omega = 0.04$, $l=4$]{\includegraphics[width=0.5\textwidth]{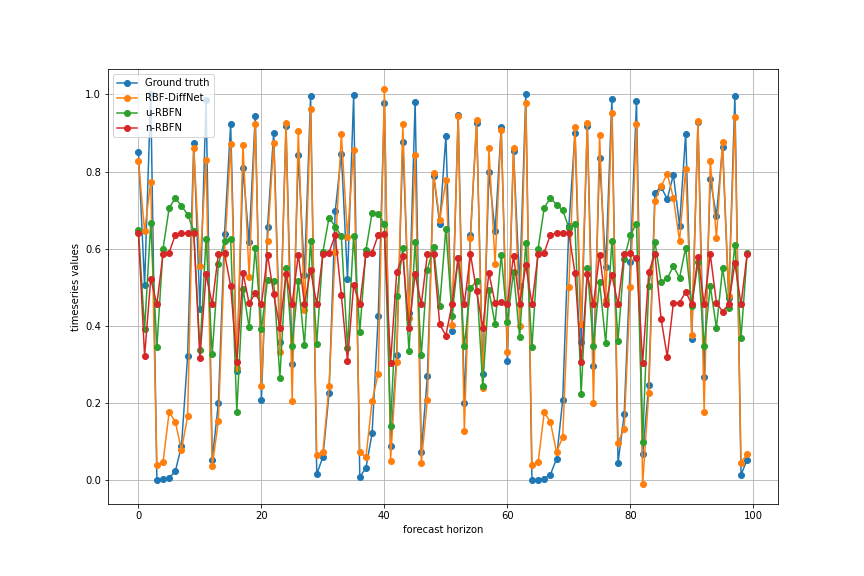}}
    \subfloat[$\omega = 0.04$, $l=8$]{\includegraphics[width=0.5\textwidth]{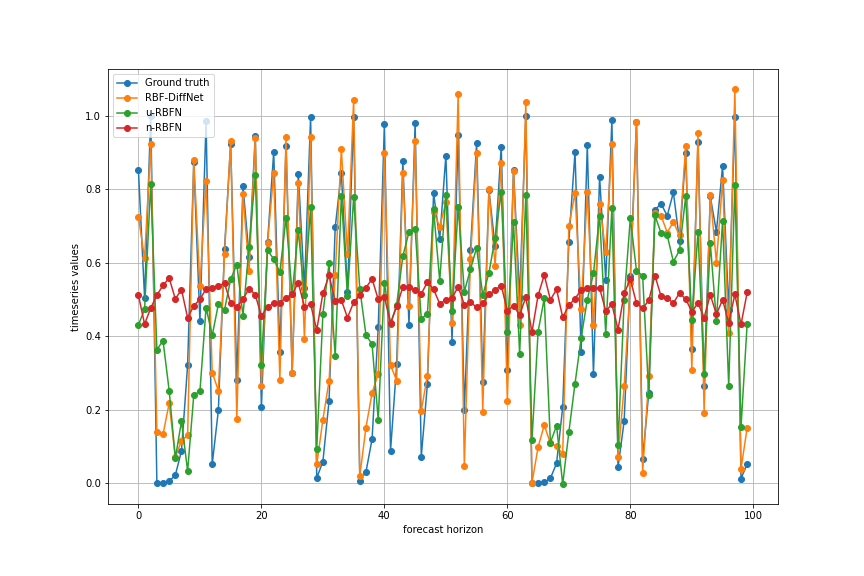}}\\
    \subfloat[$\omega = 0.04$, $l=16$]{\includegraphics[width=0.5\textwidth]{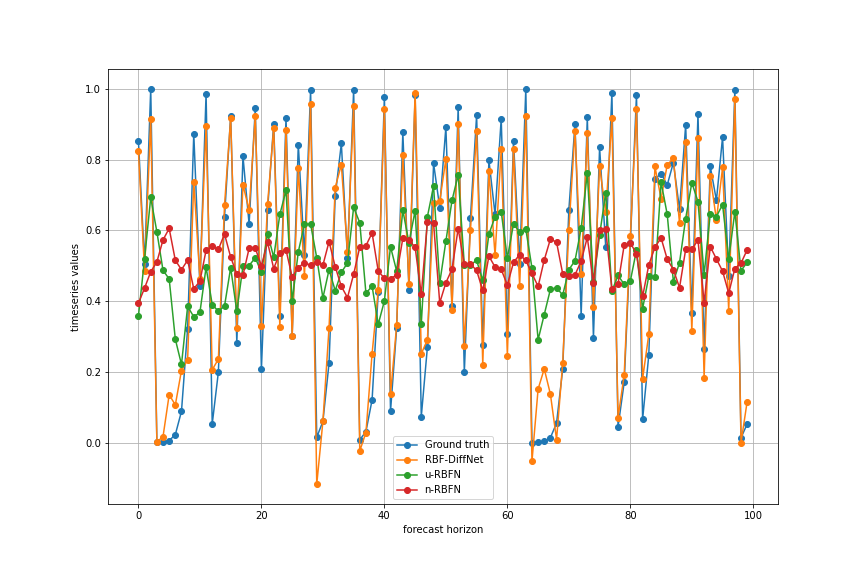}}
    \subfloat[$\omega = 0.08$, $l=1$]{\includegraphics[width=0.5\textwidth]{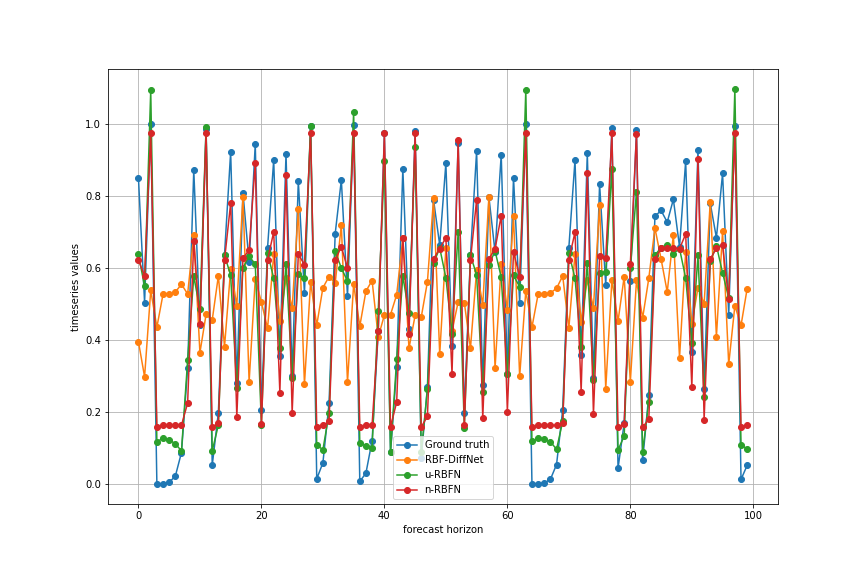}}\\
    \subfloat[$\omega = 0.08$, $l=2$]{\includegraphics[width=0.5\textwidth]{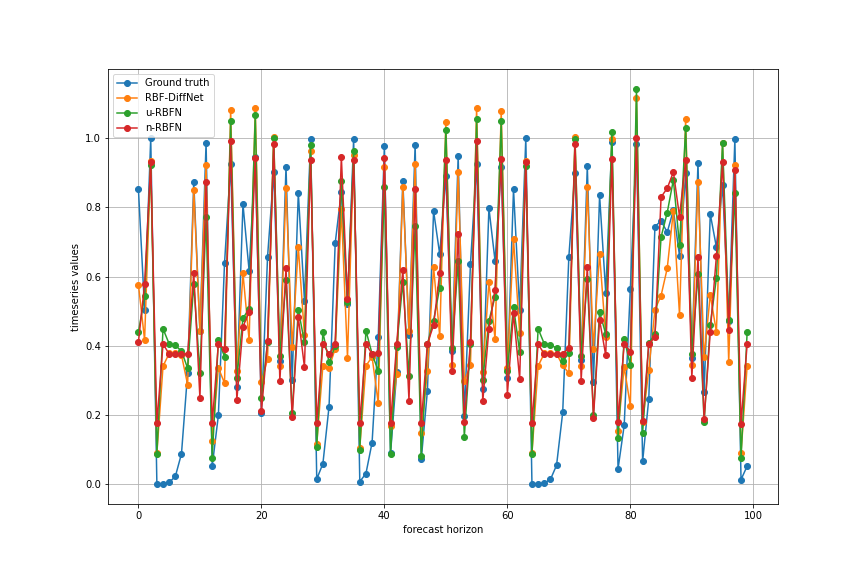}}
    \subfloat[$\omega = 0.08$, $l=4$]{\includegraphics[width=0.5\textwidth]{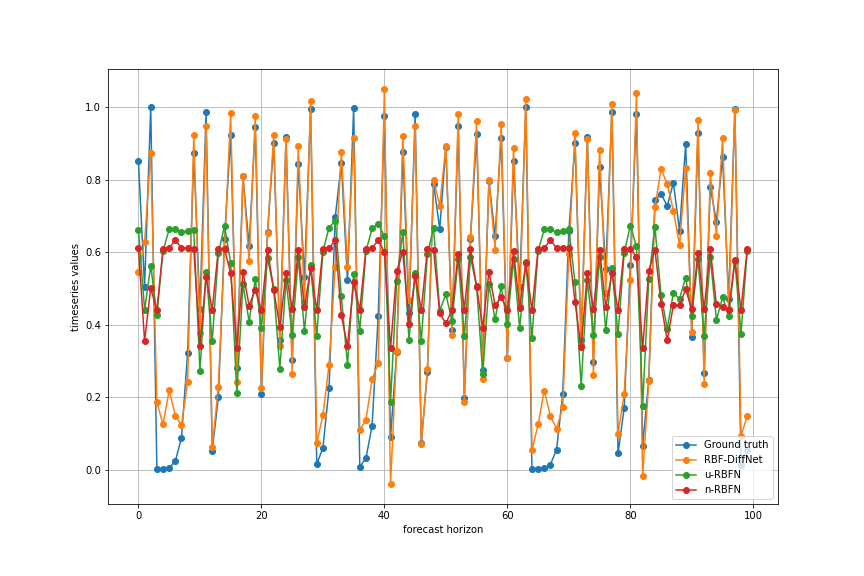}}
    \caption{Model predictions on the logistic map at different noise levels. RBF-DiffNet is the proposed differential RBF network; u-RBFN and n-RBFN are the unnormalised and normalised RBF networks respectively. $\omega$ is the noise variance; $l$ is the lookback window length.}
    \label{log_3}
\end{figure}

\begin{figure}[tbph]
    \centering
    \subfloat[$\omega = 0.08$, $l=8$]{\includegraphics[width=0.5\textwidth]{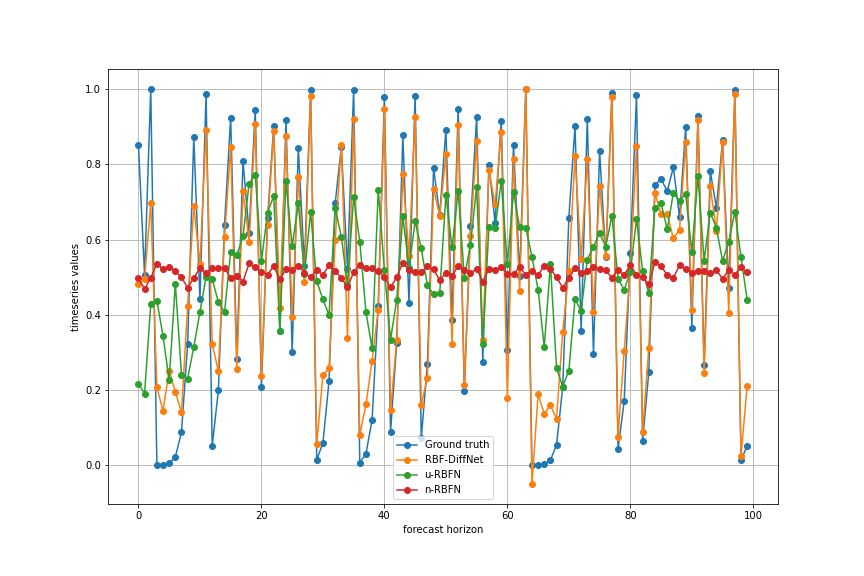}}
    \subfloat[$\omega = 0.08$, $l=16$]{\includegraphics[width=0.5\textwidth]{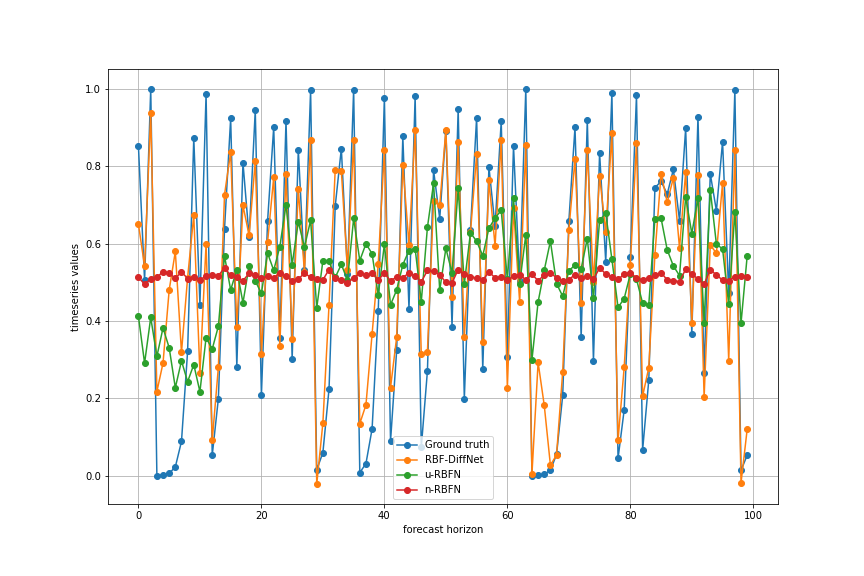}}\\
    \subfloat[$\omega = 0.12$, $l=1$]{\includegraphics[width=0.5\textwidth]{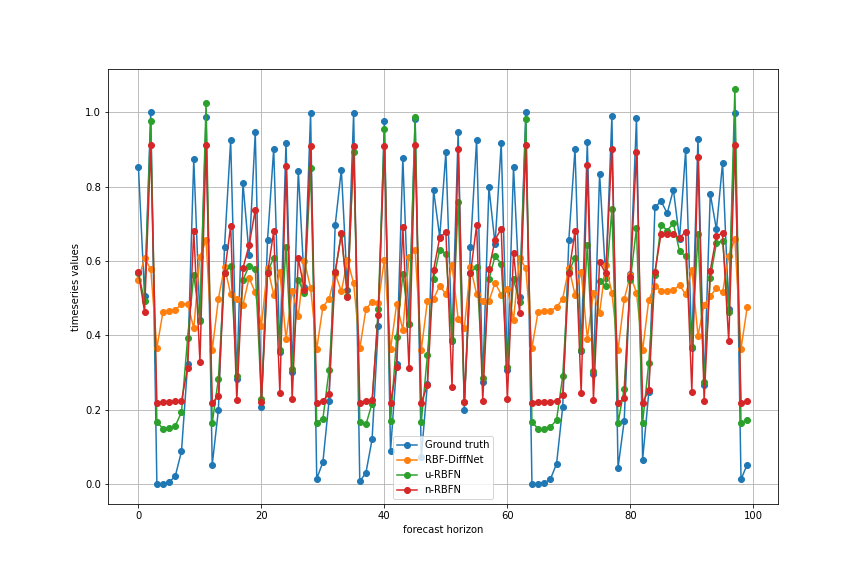}}
    \subfloat[$\omega = 0.12$, $l=2$]{\includegraphics[width=0.5\textwidth]{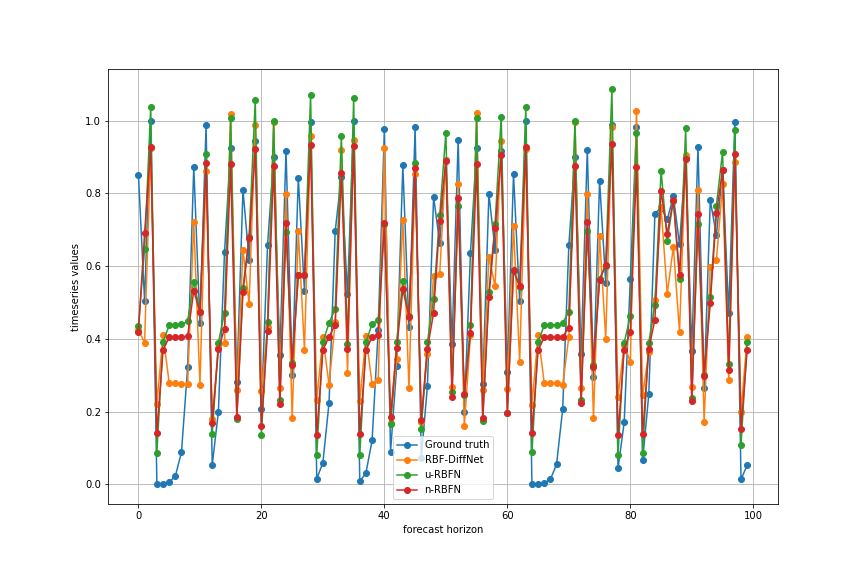}}\\
    \subfloat[$\omega = 0.12$, $l=4$]{\includegraphics[width=0.5\textwidth]{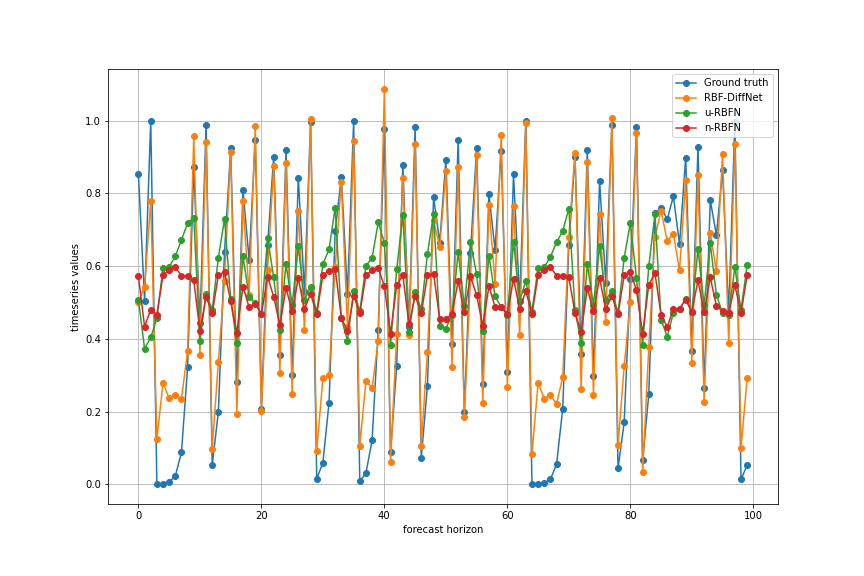}}
    \subfloat[$\omega = 0.12$, $l=8$]{\includegraphics[width=0.5\textwidth]{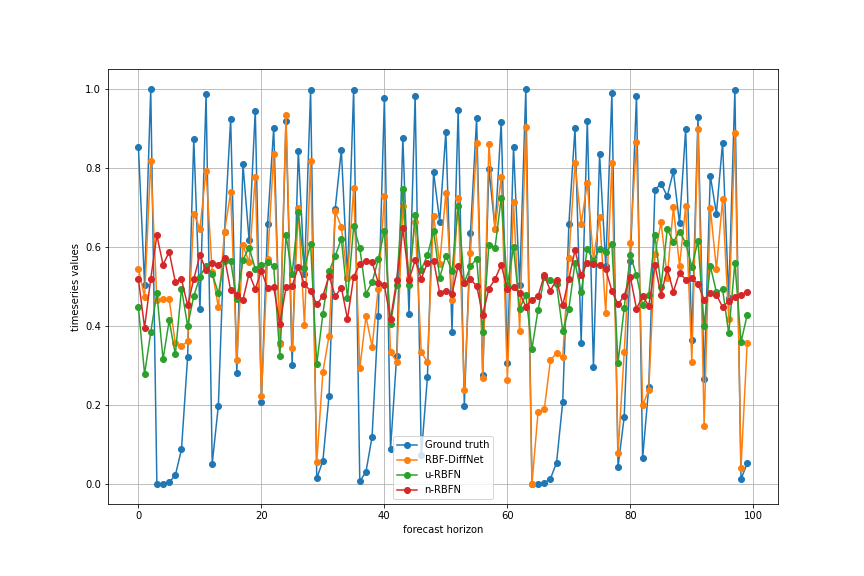}}
    \caption{Model predictions on the logistic map at different noise levels. RBF-DiffNet is the proposed differential RBF network; u-RBFN and n-RBFN are the unnormalised and normalised RBF networks respectively. $\omega$ is the noise variance; $l$ is the lookback window length.}
    \label{log_4}
\end{figure}

\begin{figure}[tbph]
    \centering
    \subfloat[$\omega = 0.12$, $l=16$]{\includegraphics[width=0.5\textwidth]{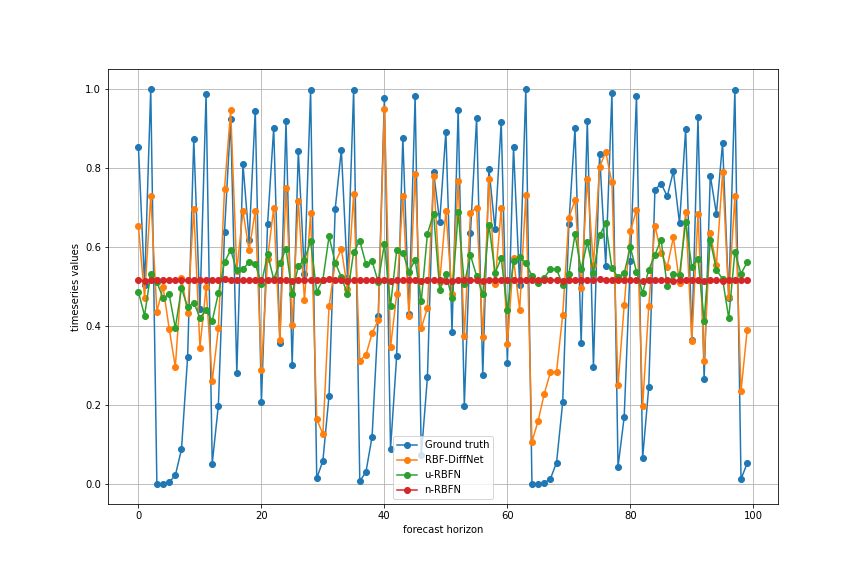}}
    \caption{Model predictions on the logistic map at different noise levels. RBF-DiffNet is the proposed differential RBF network; u-RBFN and n-RBFN are the unnormalised and normalised RBF networks respectively. $\omega$ is the noise variance; $l$ is the lookback window length.}
    \label{log_5}
\end{figure}

\section{M5 results}
This section contains graphical results from model predictions on the M5 dataset (level 8 aggregation) as described in Section \ref{m5_exp}.
\begin{figure}[tbph]
    \centering
    \subfloat[Series 1]{\includegraphics[width=0.5\textwidth]{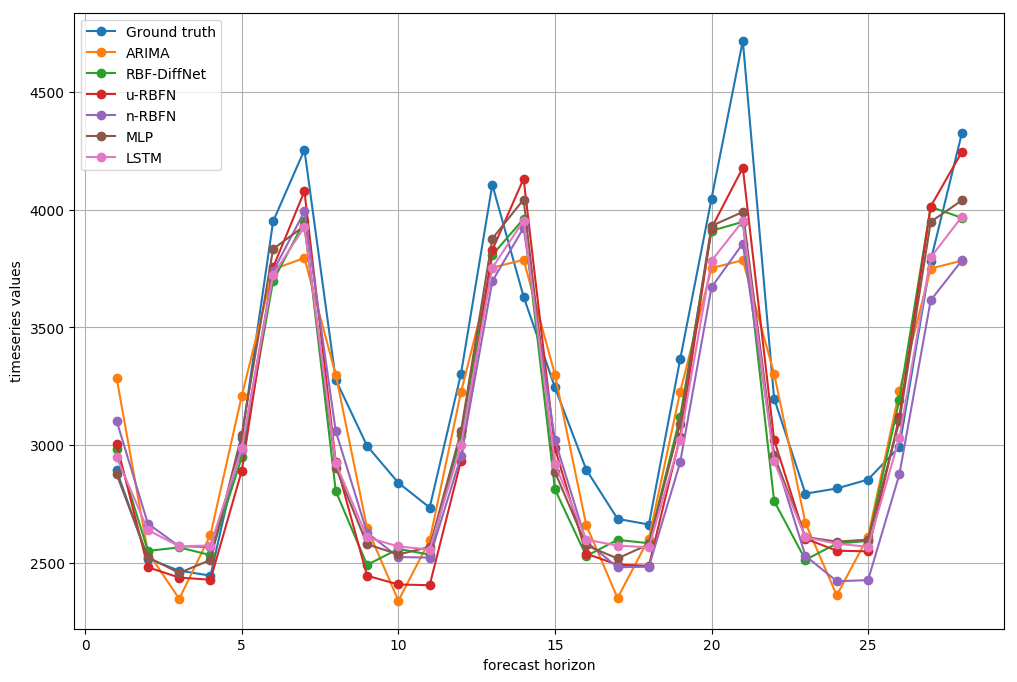}}
    \subfloat[Series 2]{\includegraphics[width=0.5\textwidth]{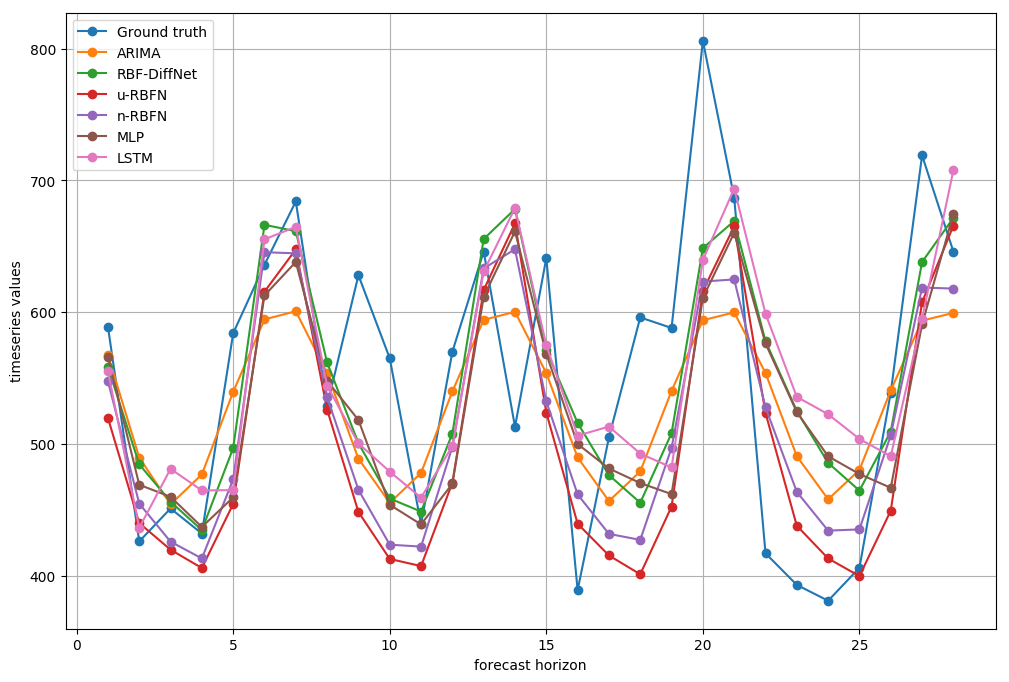}}\\
    \subfloat[Series 3]{\includegraphics[width=0.5\textwidth]{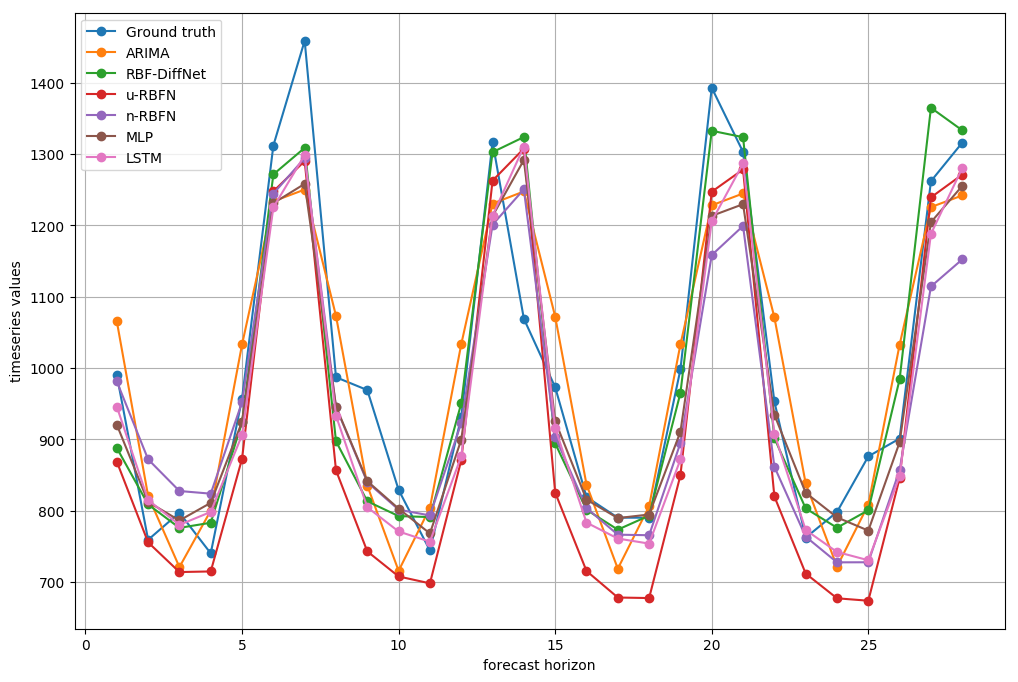}}
    \subfloat[Series 4]{\includegraphics[width=0.5\textwidth]{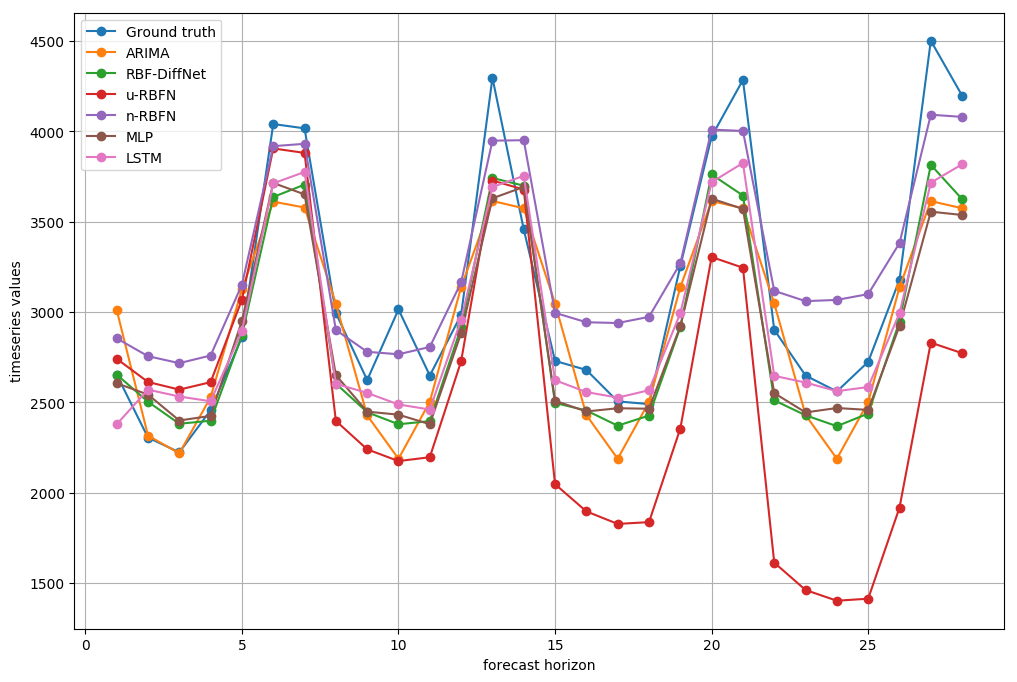}}\\
    \subfloat[Series 5]{\includegraphics[width=0.5\textwidth]{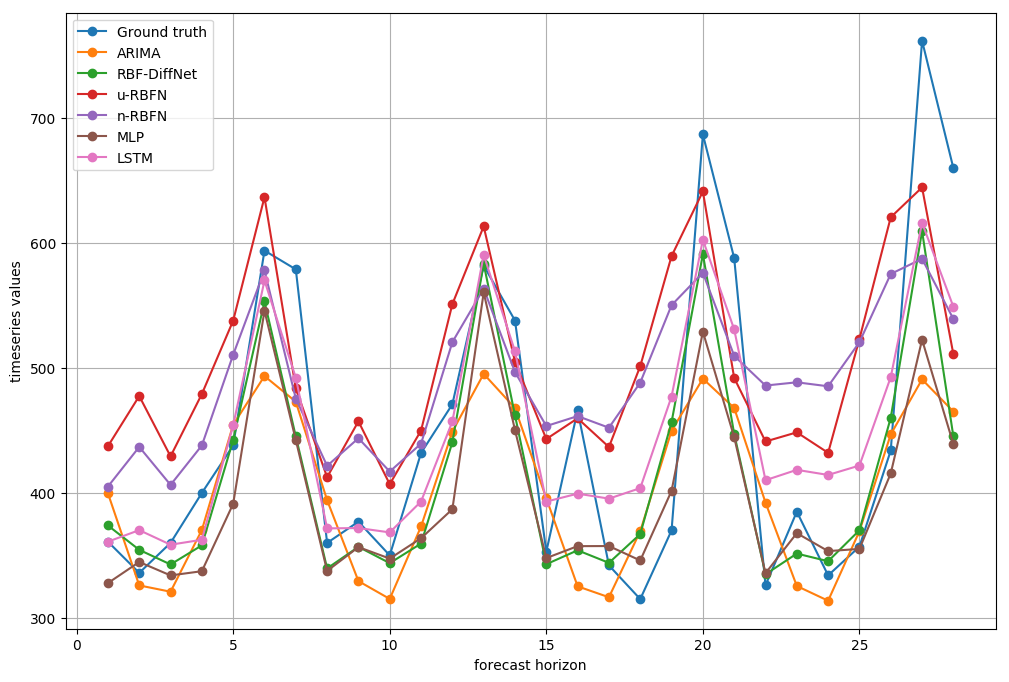}}
    \subfloat[Series 6]{\includegraphics[width=0.5\textwidth]{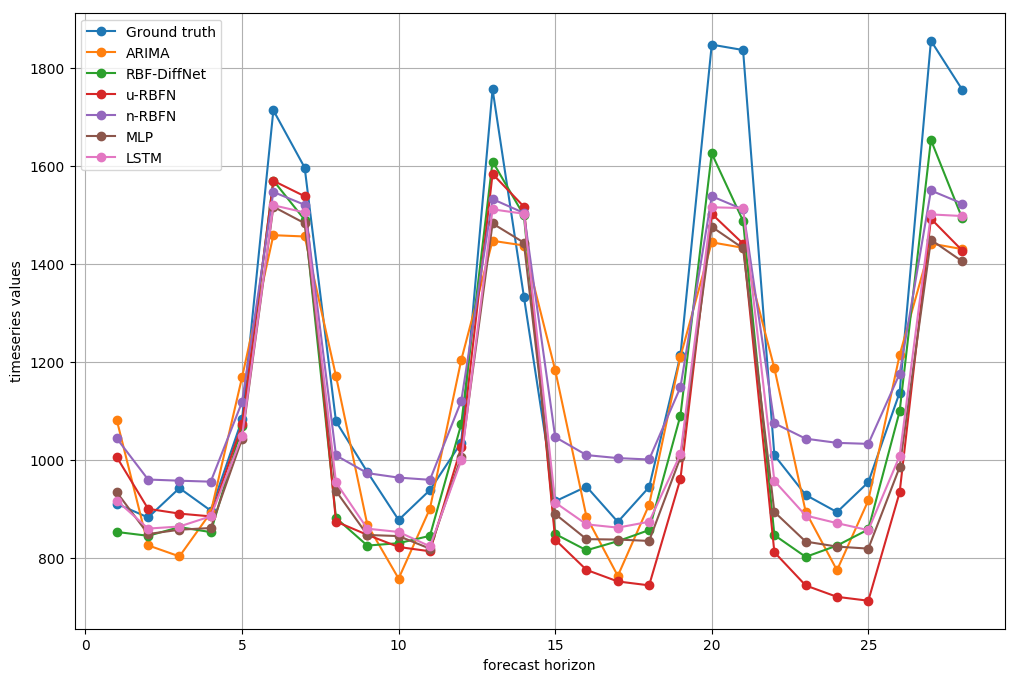}}
    \caption{Model predictions on M5 dataset (level 8 aggregation). RBF-DiffNet is the proposed differential RBF network; u-RBFN and n-RBFN are the unnormalised and normalised RBF networks respectively.}
    \label{m5_1}
\end{figure}

\begin{figure}[tbph]
    \centering
    \subfloat[Series 7]{\includegraphics[width=0.5\textwidth]{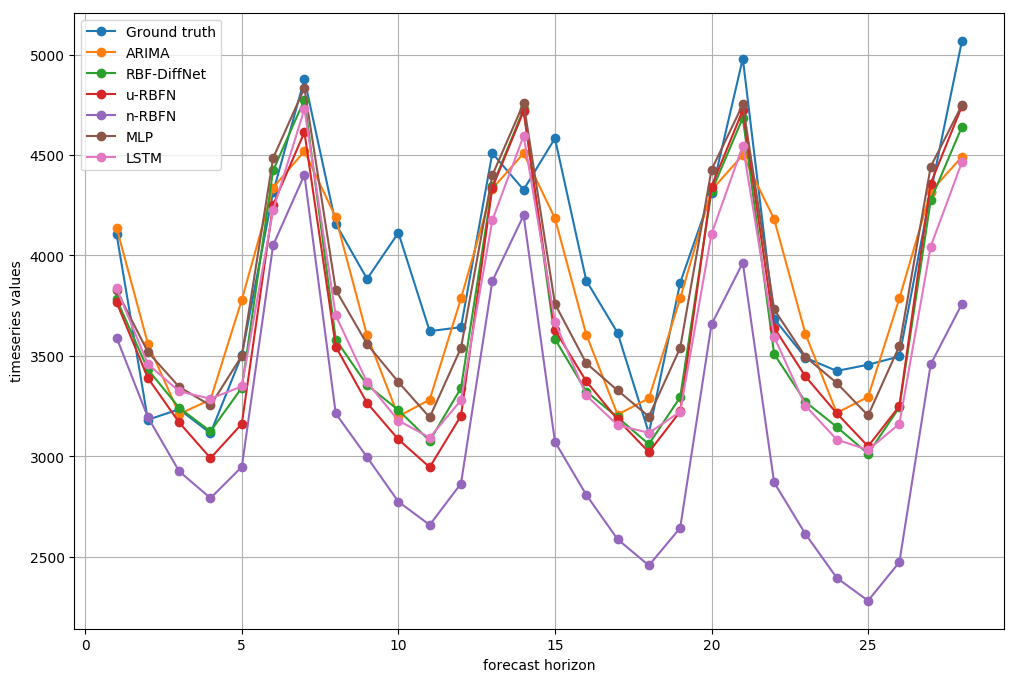}}
    \subfloat[Series 8]{\includegraphics[width=0.5\textwidth]{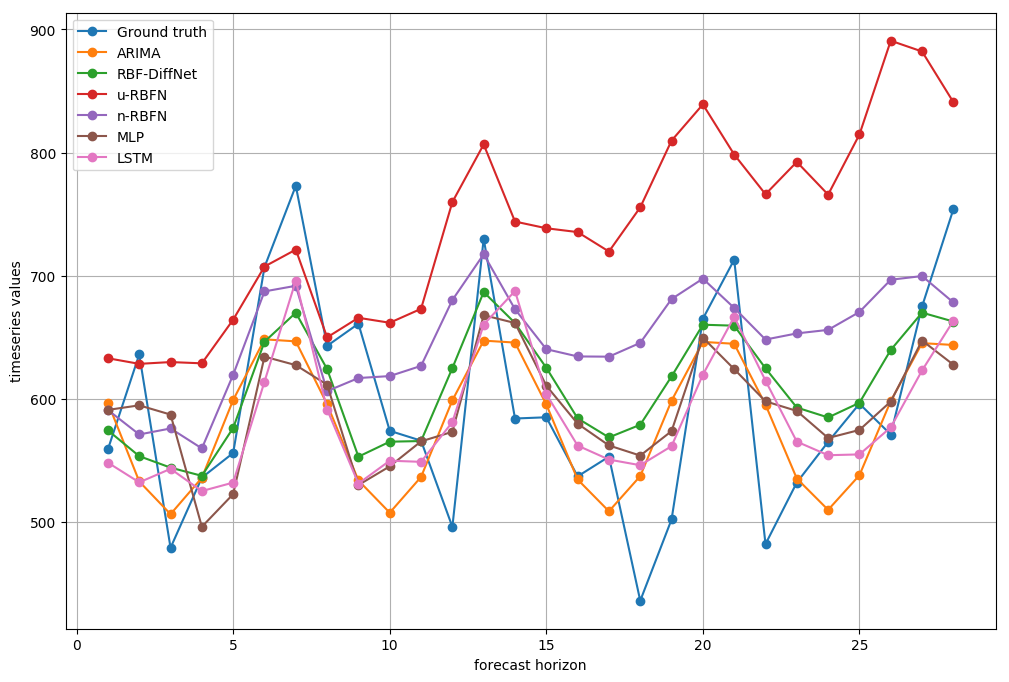}}\\
    \subfloat[Series 9]{\includegraphics[width=0.5\textwidth]{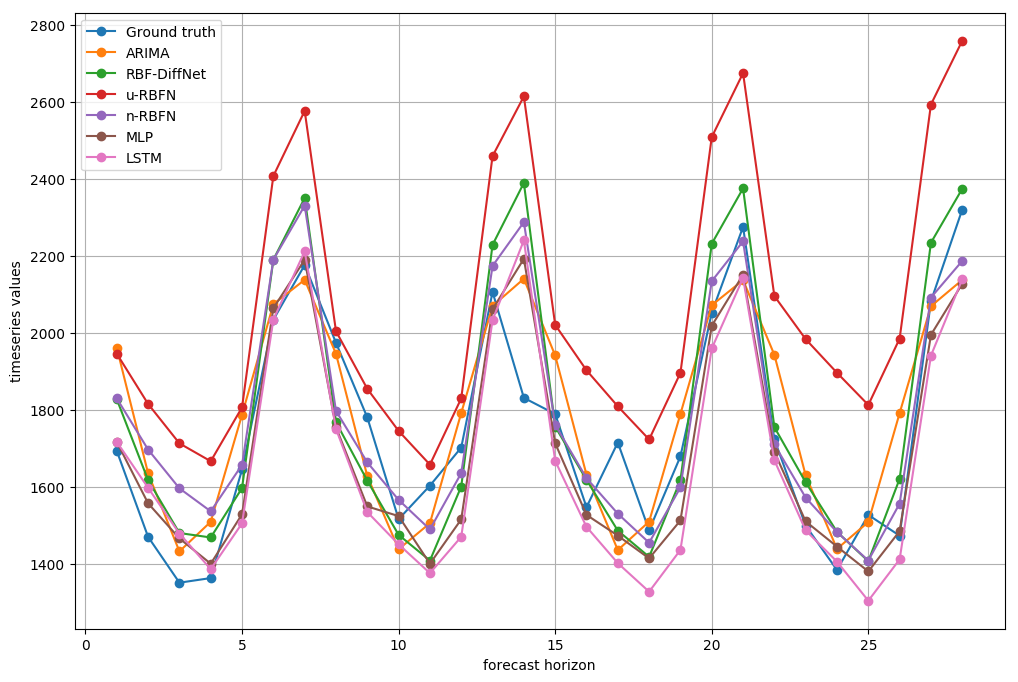}}
    \subfloat[Series 10]{\includegraphics[width=0.5\textwidth]{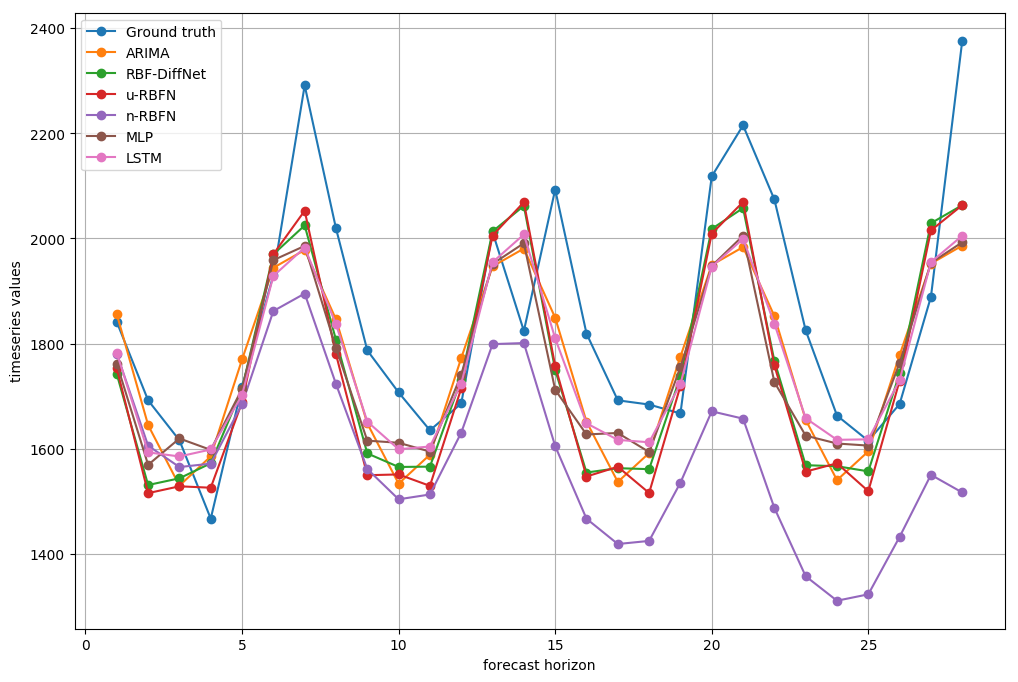}}\\
    \subfloat[Series 11]{\includegraphics[width=0.5\textwidth]{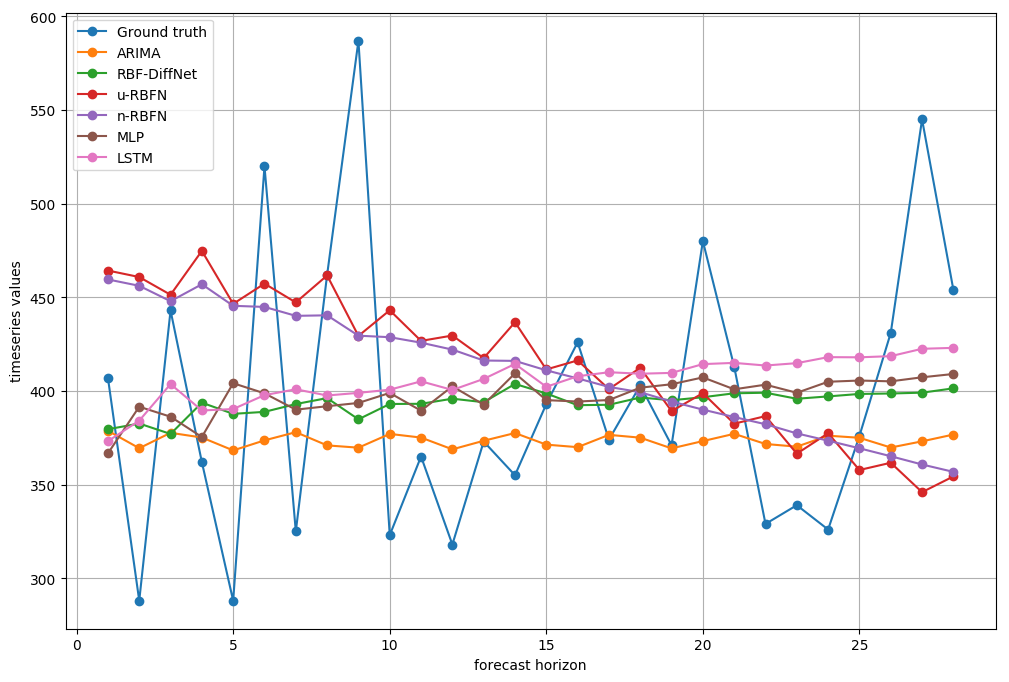}}
    \subfloat[Series 12]{\includegraphics[width=0.5\textwidth]{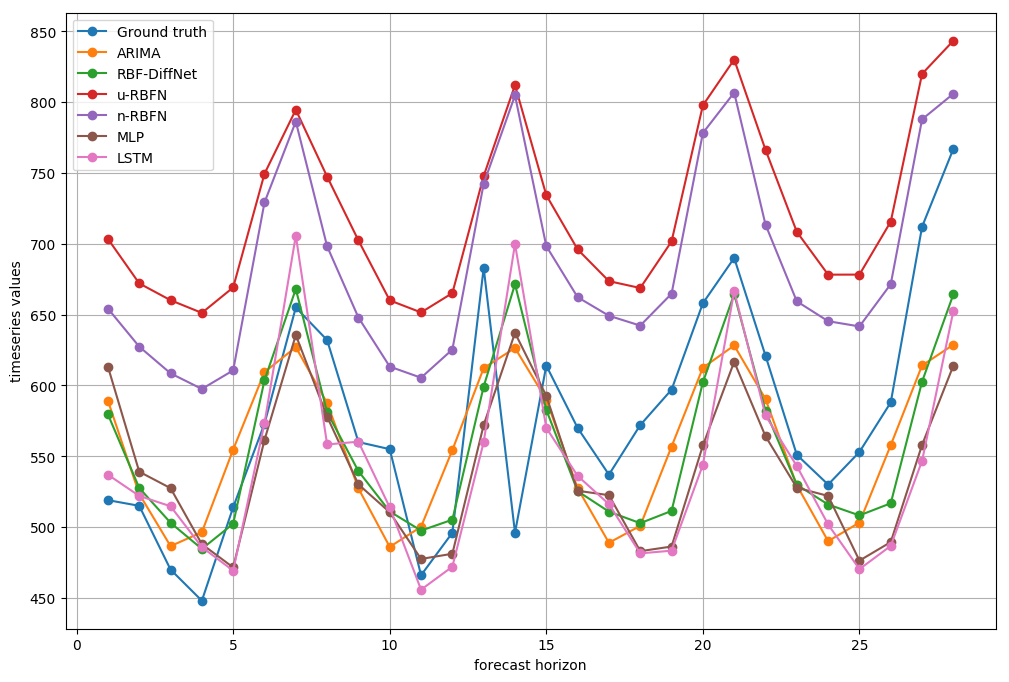}}
    \caption{Model predictions on M5 dataset (level 8 aggregation). RBF-DiffNet is the proposed differential RBF network; u-RBFN and n-RBFN are the unnormalised and normalised RBF networks respectively.}
    \label{m5_2}
\end{figure}

\begin{figure}[tbph]
    \centering
    \subfloat[Series 13]{\includegraphics[width=0.5\textwidth]{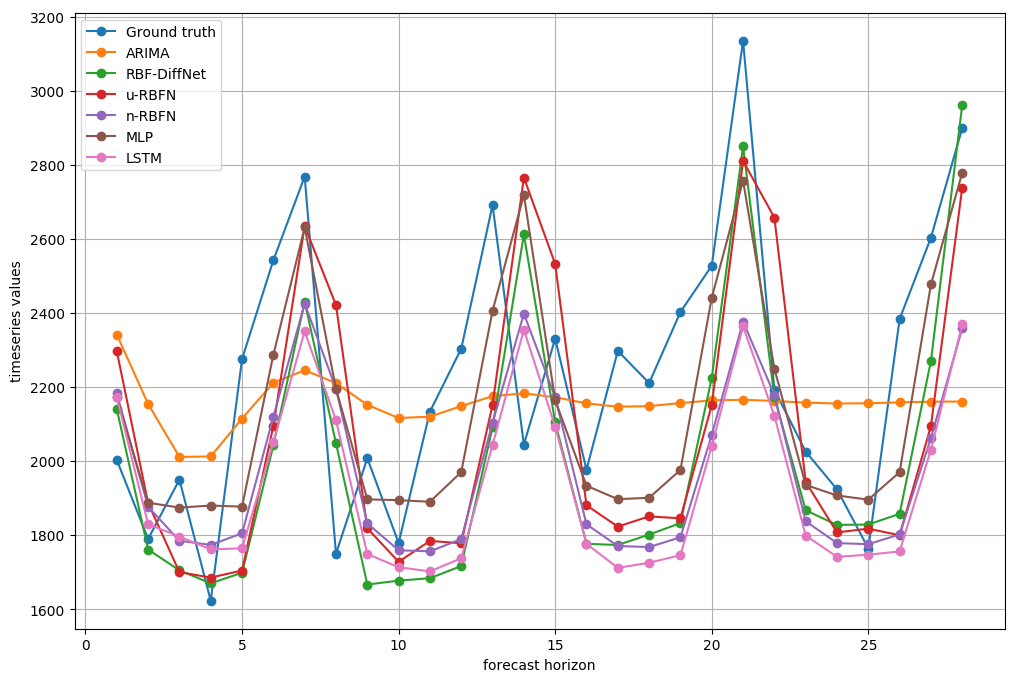}}
    \subfloat[Series 14]{\includegraphics[width=0.5\textwidth]{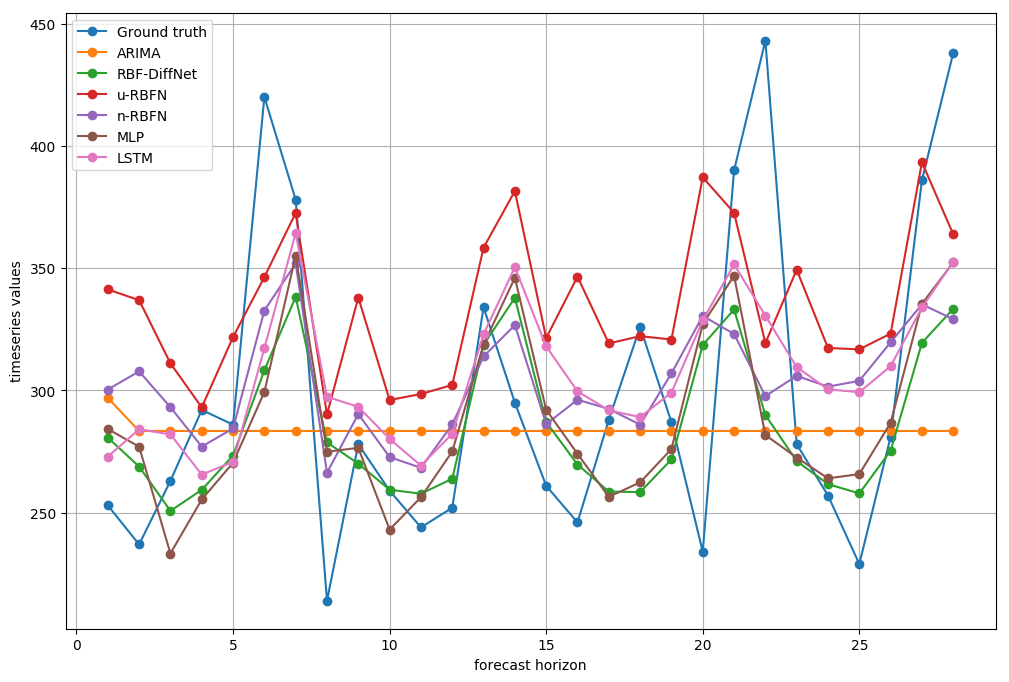}}\\
    \subfloat[Series 15]{\includegraphics[width=0.5\textwidth]{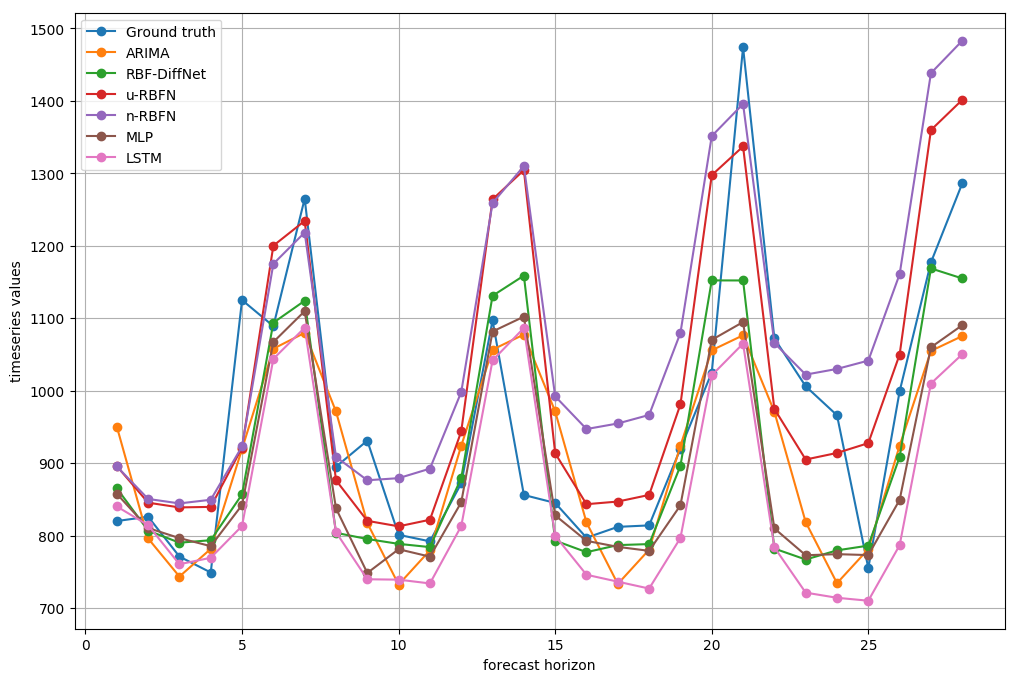}}
    \subfloat[Series 16]{\includegraphics[width=0.5\textwidth]{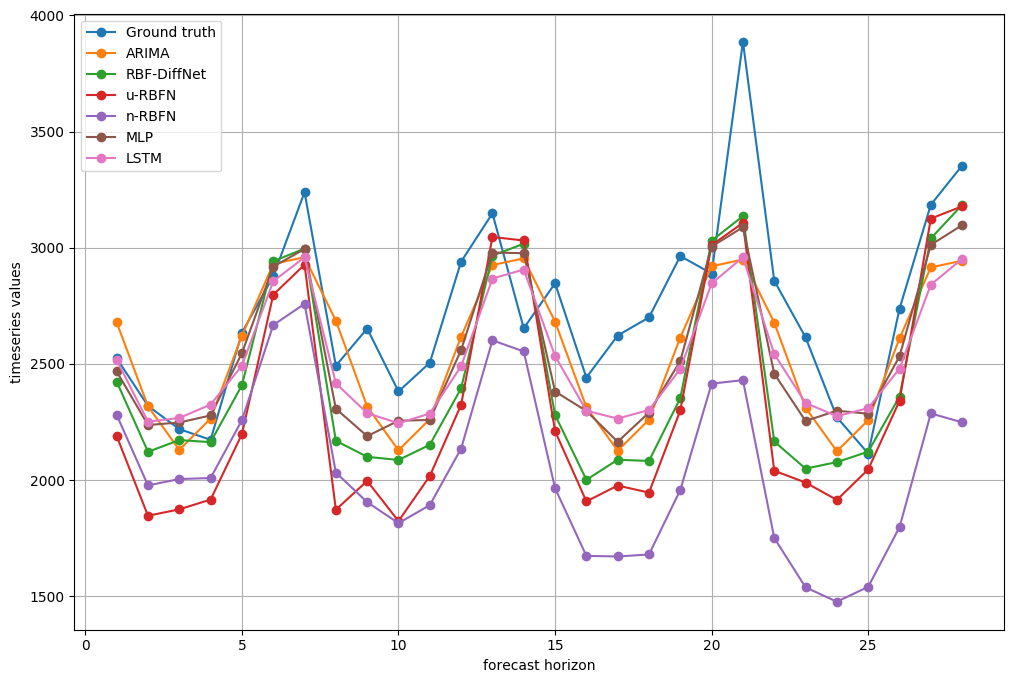}}\\
    \subfloat[Series 17]{\includegraphics[width=0.5\textwidth]{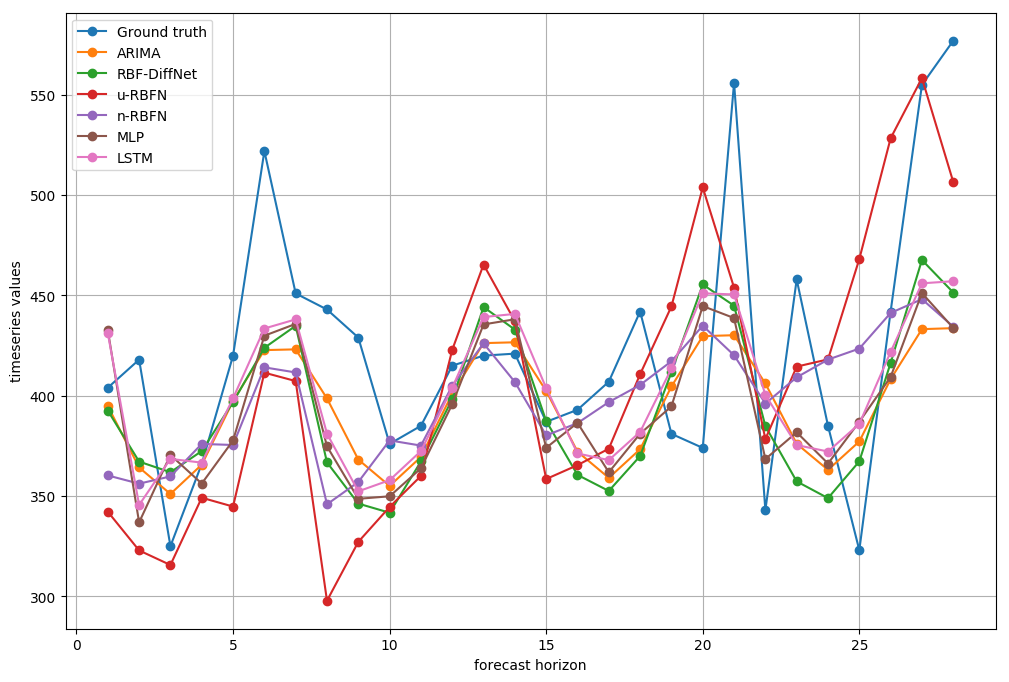}}
    \subfloat[Series 18]{\includegraphics[width=0.5\textwidth]{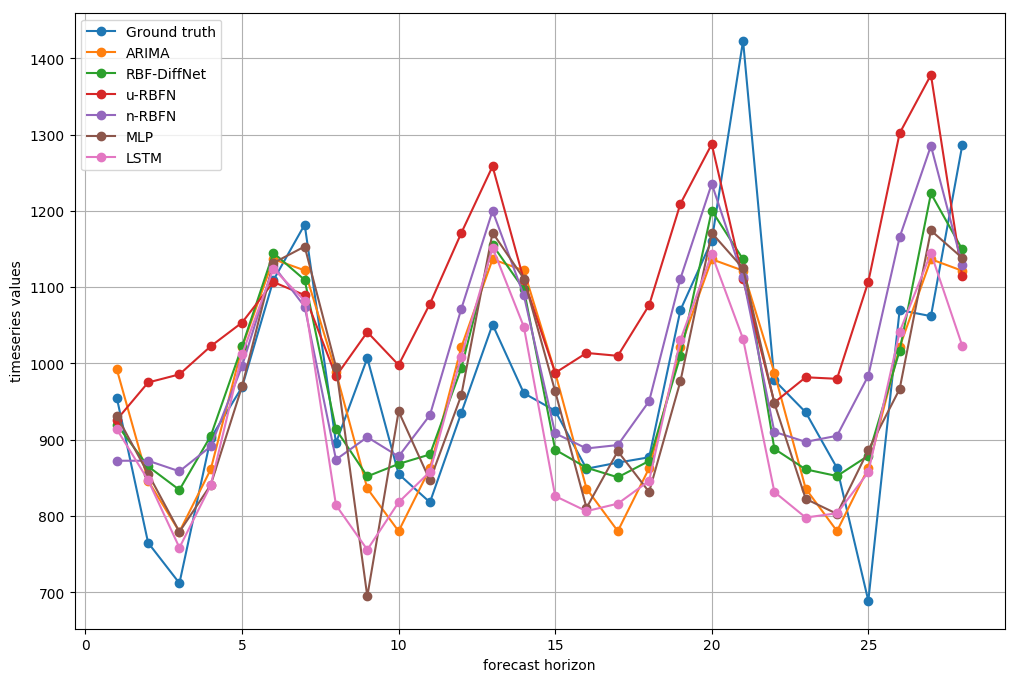}}
    \caption{Model predictions on M5 dataset (level 8 aggregation). RBF-DiffNet is the proposed differential RBF network; u-RBFN and n-RBFN are the unnormalised and normalised RBF networks respectively.}
    \label{m5_3}
\end{figure}

\begin{figure}[tbph]
    \centering
    \subfloat[Series 19]{\includegraphics[width=0.5\textwidth]{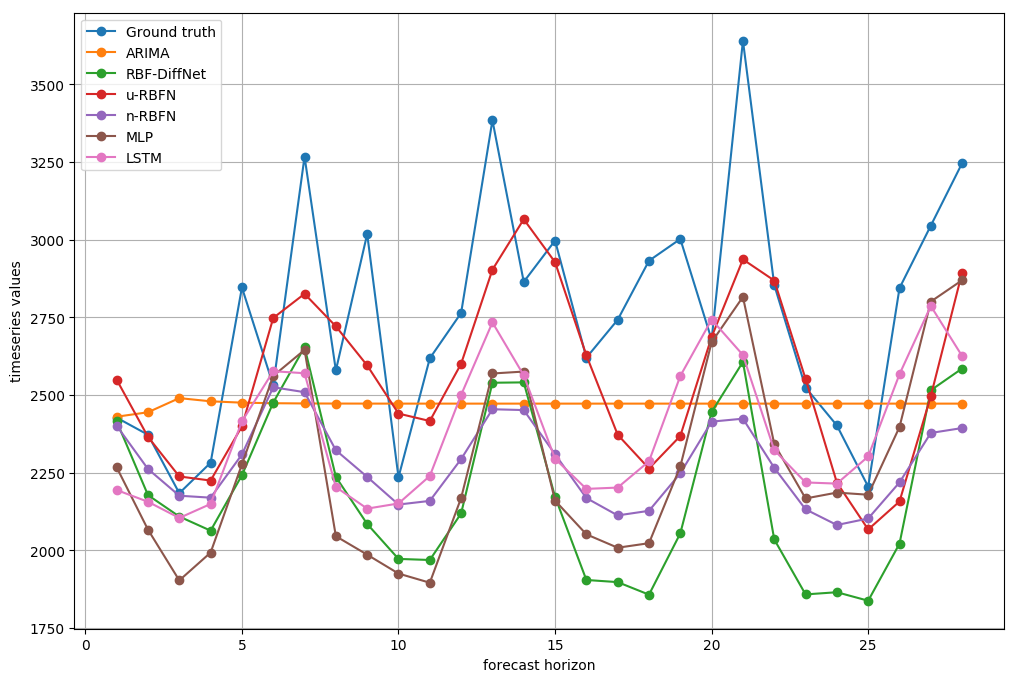}}
    \subfloat[Series 20]{\includegraphics[width=0.5\textwidth]{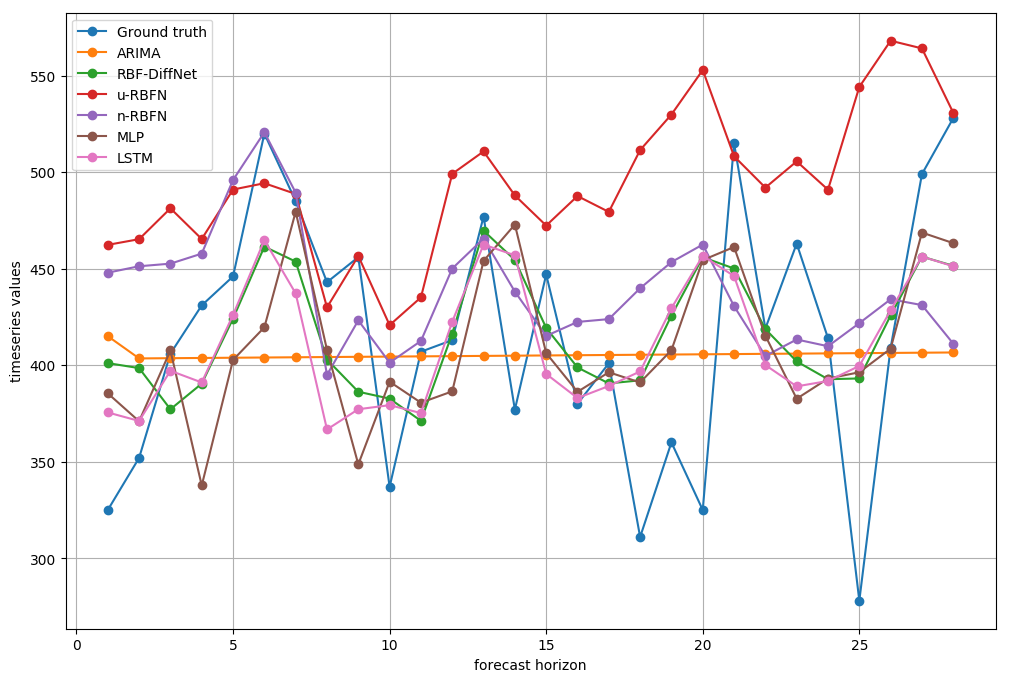}}\\
    \subfloat[Series 21]{\includegraphics[width=0.5\textwidth]{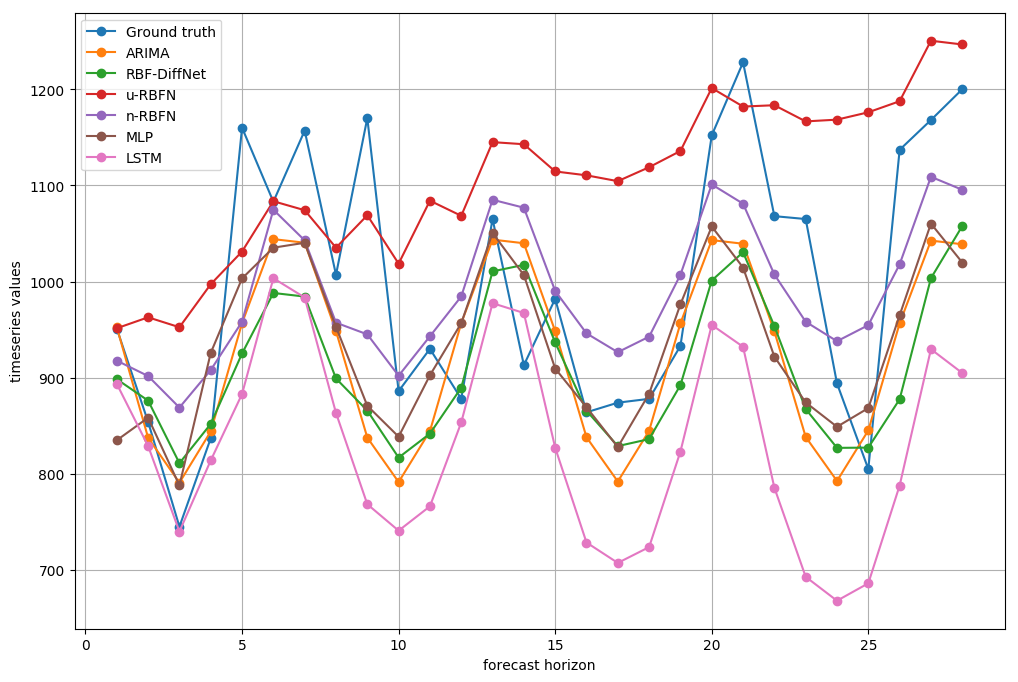}}
    \subfloat[Series 22]{\includegraphics[width=0.5\textwidth]{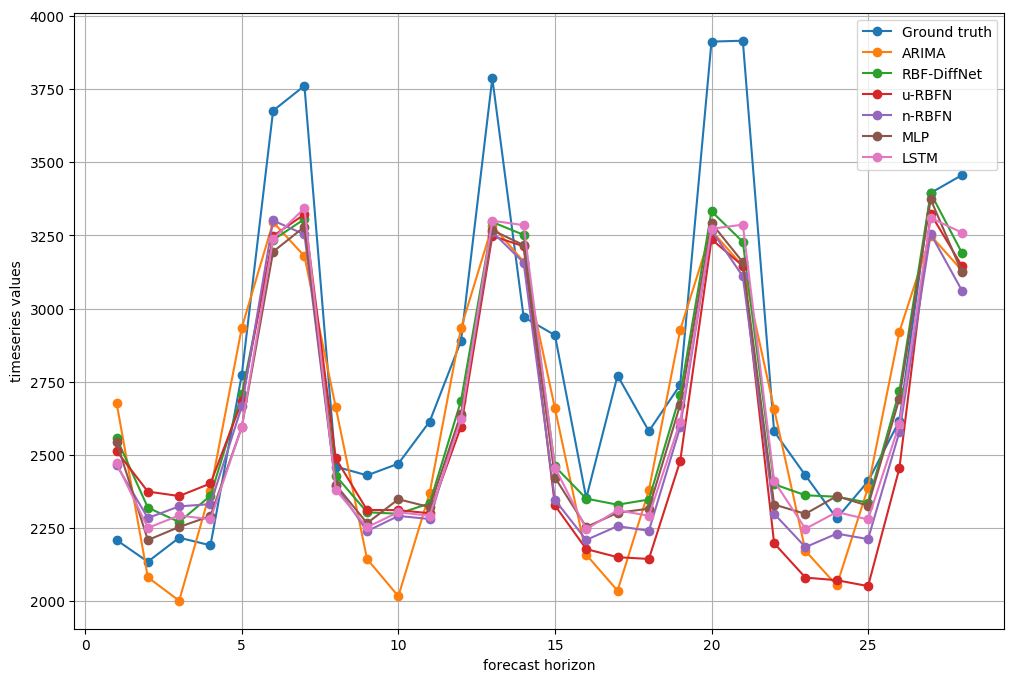}}\\
    \subfloat[Series 23]{\includegraphics[width=0.5\textwidth]{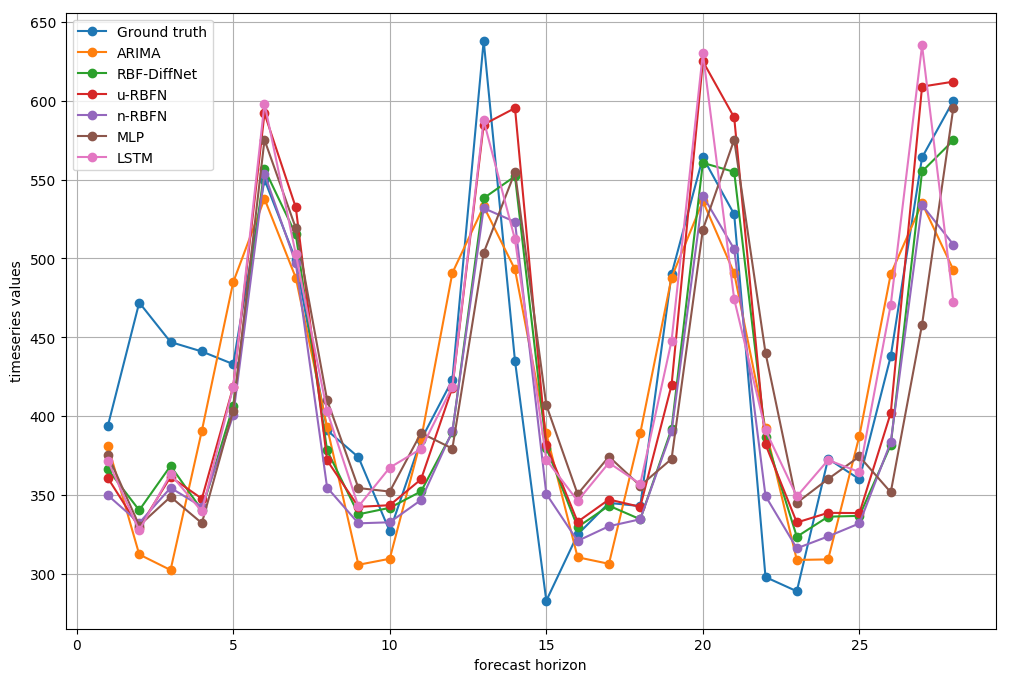}}
    \subfloat[Series 24]{\includegraphics[width=0.5\textwidth]{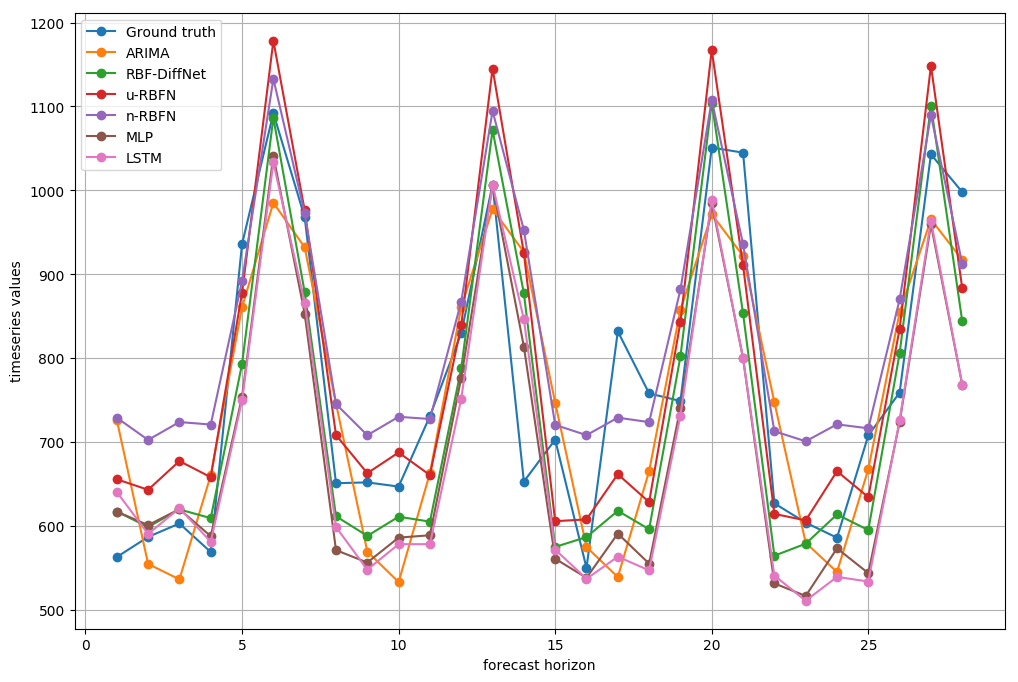}}
    \caption{Model predictions on M5 dataset (level 8 aggregation). RBF-DiffNet is the proposed differential RBF network; u-RBFN and n-RBFN are the unnormalised and normalised RBF networks respectively.}
    \label{m5_4}
\end{figure}

\begin{figure}[tbph]
    \centering
    \subfloat[Series 25]{\includegraphics[width=0.5\textwidth]{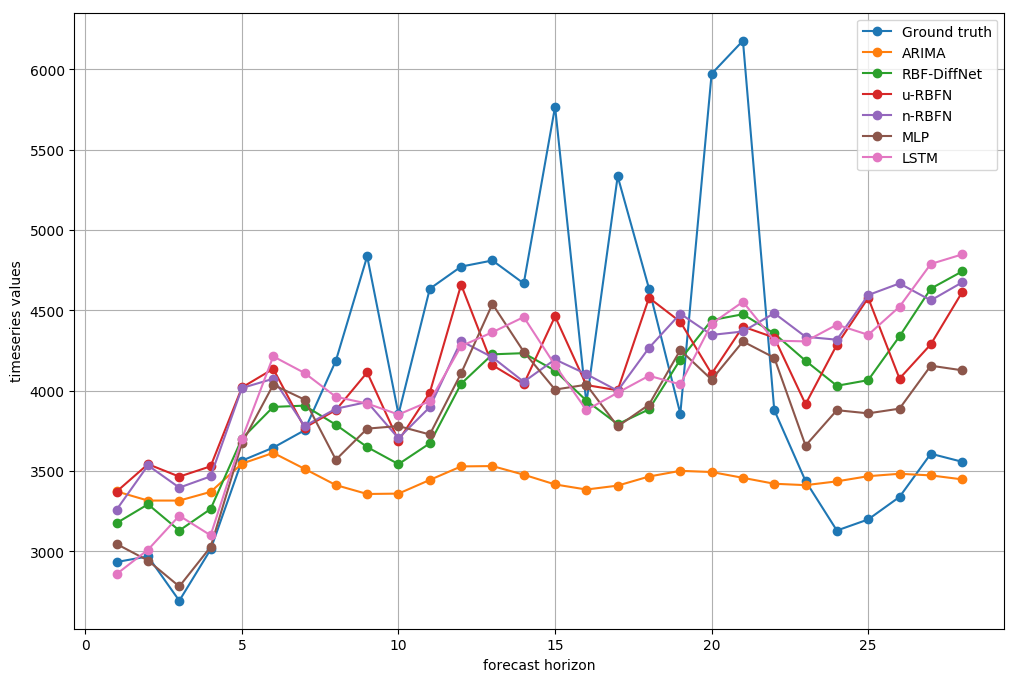}}
    \subfloat[Series 26]{\includegraphics[width=0.5\textwidth]{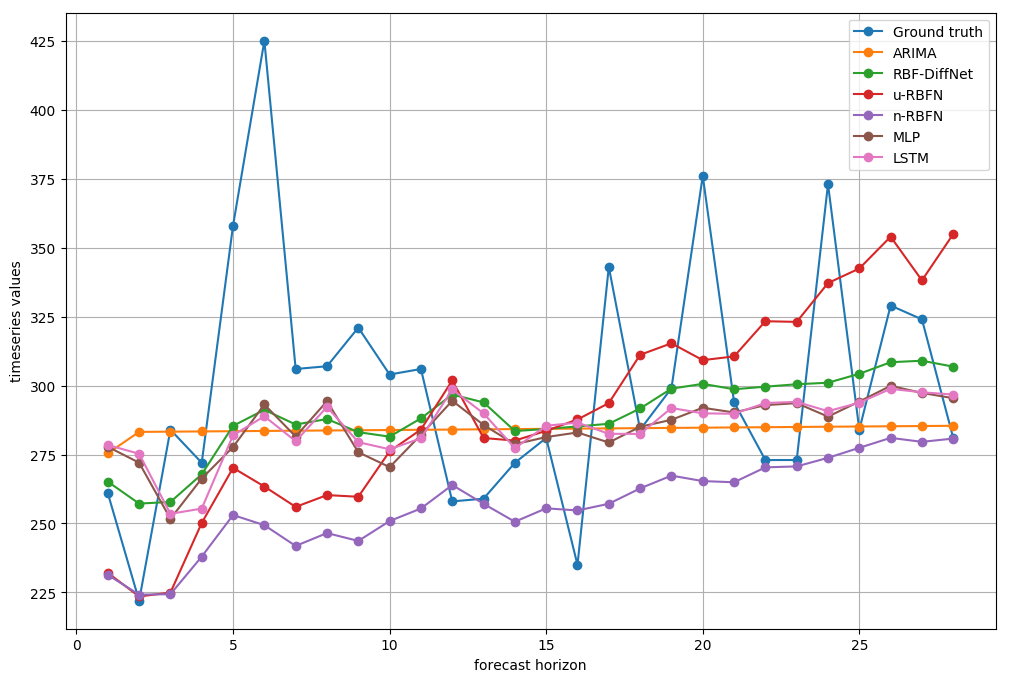}}\\
    \subfloat[Series 27]{\includegraphics[width=0.5\textwidth]{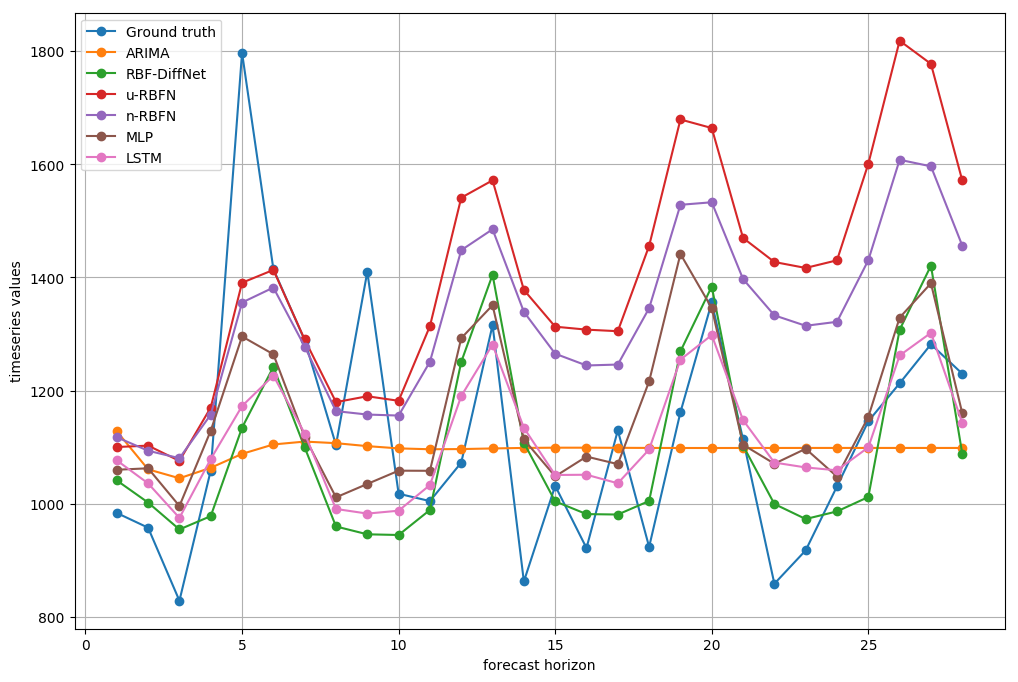}}
    \subfloat[Series 28]{\includegraphics[width=0.5\textwidth]{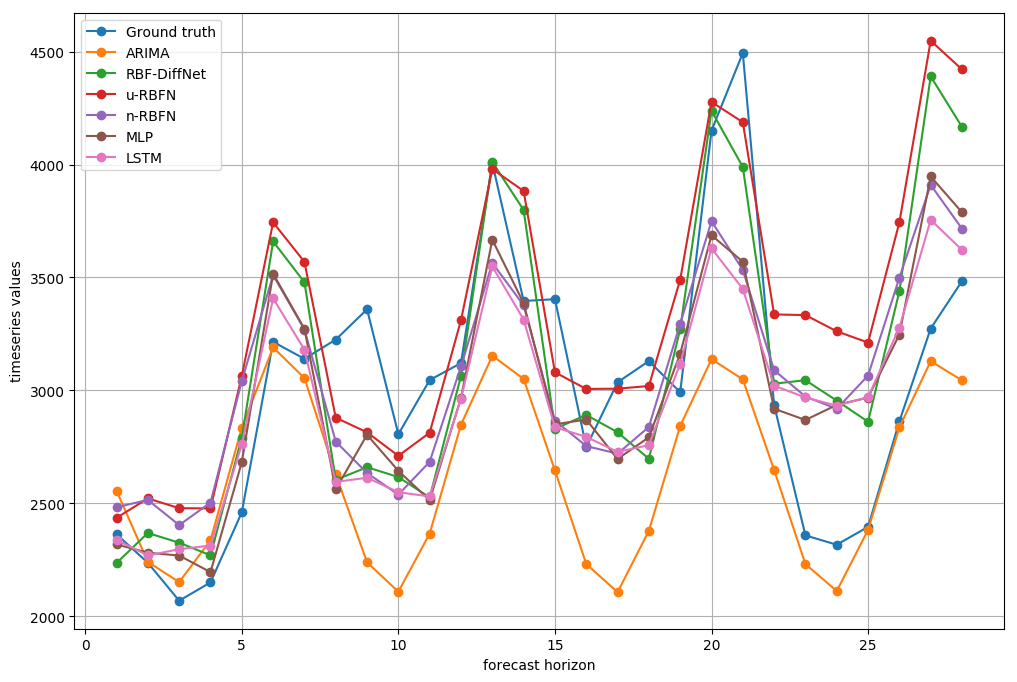}}\\
    \subfloat[Series 29]{\includegraphics[width=0.5\textwidth]{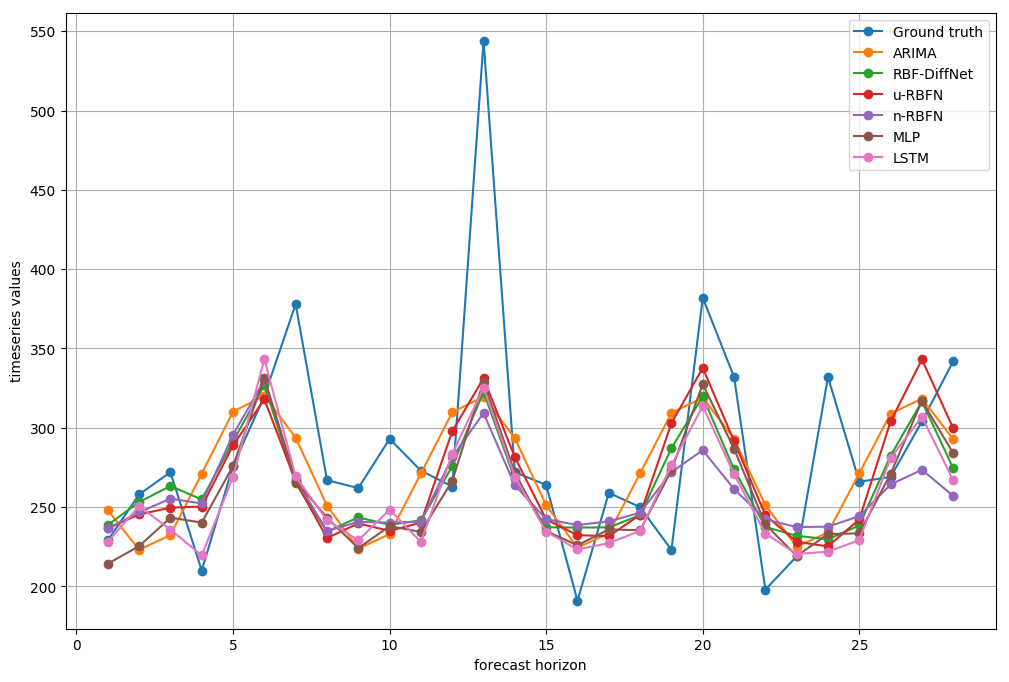}}
    \subfloat[Series 30]{\includegraphics[width=0.5\textwidth]{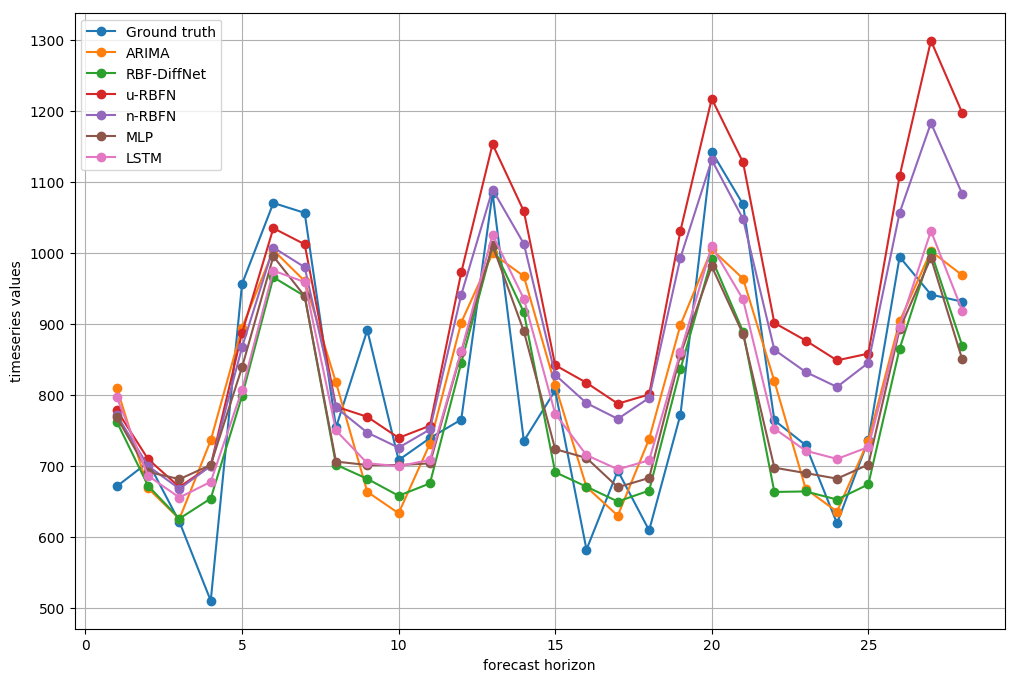}}
    \caption{Model predictions on M5 dataset (level 8 aggregation). RBF-DiffNet is the proposed differential RBF network; u-RBFN and n-RBFN are the unnormalised and normalised RBF networks respectively.}
    \label{m5_5}
\end{figure}

\end{document}